\documentclass[12pt]{article}

\usepackage[utf8]{inputenc} 
\usepackage{csquotes}
\usepackage[T1]{fontenc}    
\usepackage{url}            
\usepackage{booktabs}       
\usepackage{nicefrac}       
\usepackage{lipsum}
\usepackage[english]{babel} 
\usepackage{amsmath}
\usepackage{amsfonts}
\usepackage{amsthm} 
\usepackage{amssymb}
\usepackage[hang, small, labelfont=bf, up]{caption} 
\usepackage{lastpage}
\usepackage{graphicx} 
\usepackage{enumitem} 
\usepackage[capitalise]{cleveref}
\usepackage{comment}
\usepackage{xcolor}
\usepackage[margin=1.0in]{geometry}
\usepackage{authblk}
\usepackage{times}
\usepackage{afterpage}
\usepackage{changepage}
\usepackage{caption}
\usepackage{algorithm}
\usepackage{algpseudocode}
\usepackage[hang,flushmargin]{footmisc}
\usepackage{caption}
\usepackage{subcaption}
\usepackage{charter}
\usepackage{multirow}
\usepackage{array}
\newcommand{\PreserveBackslash}[1]{\let\temp=\\#1\let\\=\temp}
\newcolumntype{C}[1]{>{\PreserveBackslash\centering}p{#1}}

\usepackage[backend=bibtex8,
            natbib=true,
            sorting=none,
            citestyle=numeric-comp]{biblatex}          
\newcommand{\ds}        {\displaystyle}
\newcommand{\bd}[1]     { \mathbf{#1} }
\newcommand{\tr}        { \mbox{\!\! \tiny T} }  

\newcommand{\dblvert}   { \left| \left|  }
\newcommand{\dbrvert}   { \right| \right| }
\newcommand{\norm}[1]   { \dblvert #1 \dbrvert_2 }

\newcommand{\fnorm}[1]  { 
    \dblvert #1 \dbrvert_{ \mbox{\scriptsize F} } 
}

\newcommand{\quadand}   { \quad \mbox{and} \quad }
\newcommand{\matvec}[2] { 
    \left[ \begin{array}{#1} #2 \end{array} \right] 
} 
\newcommand{\Xtune}     {  \bd{X}_{\mbox{\texttt{\scriptsize t}}} }
\newcommand{\atune}     {  \bd{a}_{\mbox{\texttt{\scriptsize t}}} }
\newcommand{\ytune}     {  \bd{y}_{\mbox{\texttt{\scriptsize t}}} }
\newcommand{\estav}     { \widehat{\bd{a}}_2 }

\newcommand{\Xcorr}[1]  { \bd{X}_{#1}^{ \mbox{\tiny BC} } }

\newcommand{\voc}[1]    { \bd{v}_{\mbox{\tiny $\varnothing$,#1}} }
\newcommand{\Vnot}      { \bd{V}_{\! \! \mbox{\tiny $\varnothing$}} }
\newcommand{\Unot}      { \bd{U}_{\! \! \mbox{\tiny $\varnothing$}} }
\newcommand{\SG}        {\texttt{SG}}
\newcommand{\EILERS}    {\texttt{EILERS}}
\newcommand{\EMSC}      {\texttt{EMSC}}
\newcommand{\SPBC}      {\texttt{SPBC}}
\newcommand{\SPBCN}     {\texttt{SPBCN}}
\newcommand{\SPBCI}     {\texttt{SPBCI}}
\newcommand{\ASLS}      {\texttt{ASLS}}
\newcommand{\AIRPLS}    {\texttt{AIRPLS}}
\newcommand{\ARPLS}     {\texttt{ARPLS}}
\newcommand{\NONE}      {\texttt{NONE}}
\newcommand{\SPBCNF}    {\texttt{SPBCN:F}}
\newcommand{\SPBCNP}    {\texttt{SPBCN:P}}
\newcommand{\SPBCIF}    {\texttt{SPBCI:F}}
\newcommand{\SPBCIP}    {\texttt{SPBCI:P}}

\providecommand{\keywords}[1]
{\small	 \textbf{\textit{Keywords---}} #1  }
\title{
    {\Large \textbf{Supervised and Penalized Baseline Correction} }
}
\author[1,2]{ {\large
    Erik Andries 
    \thanks{Corresponding author: erik.andries@gmail.com} }
}
\author[3]{ {\large
    Ramin Nikzad-Langerodi
    \thanks{ramin.nikzad-langerodi@scch.at} }
} 
\affil[1]{ {\small
    \emph{Center for Advanced Research Computing} 
    \protect\\ 
    \vspace*{-0.25cm}
    University of New Mexico, Albuquerque, NM, USA}
    \protect\vspace{0.3cm}
}
\affil[2]{ {\small
    Central New Mexico Community College, Albuquerque, NM, USA}
    \protect\vspace{0.3cm}
}
\affil[3]{ {\small
    \emph{Data Science Group} 
    \protect\\
    \vspace*{-0.25cm}
    Software Competence Center Hagenberg (SCCH) GmbH, 
    Hagenberg, Austria}
}
\date{}

\begin{document}
\maketitle
\begin{abstract}
Spectroscopic measurements can show distorted spectral shapes arising 
from a mixture of absorbing and scattering contributions. These 
distortions (or baselines) often manifest themselves as non-constant 
offsets  or low-frequency oscillations.  As a result, these baselines 
can adversely affect analytical and quantitative results.  Baseline 
correction is an umbrella term where one applies pre-processing 
methods to obtain baseline spectra (the unwanted distortions) and 
then remove the distortions by differencing. However, current 
state-of-the art baseline correction methods do not utilize analyte 
concentrations even if they are available, or even if they 
contribute significantly to the observed spectral variability.  We 
examine a class of state-of-the-art methods (penalized baseline 
correction) and modify them such that they can accommodate a priori
analyte concentrations such that prediction can be enhanced.  
Performance will be assessed on two near infra-red data sets across 
both classical penalized baseline correction methods (without 
analyte information) and modified penalized baseline correction 
methods (leveraging analyte information).
\end{abstract}

%
\section{Introduction}
%
Spectroscopic measurements, e.g., those obtained from near infrared 
(NIR) instrumentation, often show distorted spectral shapes arising 
from a mixture of absorbing and scattering contributions.  NIR 
spectral scattering is caused by differences in path length due to 
physical artifacts where light ballistically deviates from a straight 
line into one or multiple paths with no absorption.  Spectrally, this 
scattering typically manifests itself as undulating alterations, 
i.e., non-constant offsets and low frequency curves; see \cite{dazzi} 
for a catalogue of spectral distortions due to scattering.  These 
scattering distortions can adversely affect qualitative or 
quantitative analytical results.  The phrase \emph{baseline 
correction} refers to pre-processing methods that 
remove the physical artifacts in spectra due to scattering.
As a consequence of baseline removal, subsequent chemical 
interpretation and quantitative analyses is then more 
valid and applicable.


Historically, a common method for baseline correction is to 
fit a quadratic or higher-order polynomial function to each 
spectrum and then use the difference between the spectrum and the 
fitted function as the corrected spectrum
\cite{ruckstuhl,schecter,mazet,morhac}.
For example, Multiplicative Scatter Correction (MSC) is 
one such procedure: it corrects each measured spectrum 
using fitted coefficients (slope and intercept) of a reference 
spectrum \cite{msc}.  (The reference spectrum is usually just 
the average spectrum of the calibration set.)  There are 
extensions to MSC (e.g., Extended MSC) that include 
first-order and/or second‐order polynomial fitting to the 
reference spectrum and wavelength axis \cite{emsc,mancini}.


Alternatively, baseline removal can also be achieved via 
derivative spectra (i.e., a scaled version of the first or 
second derivative of the original spectra). Differentiation 
removes low-frequency components (e.g., the 
second derivative removes constant and linear baselines).  
However, differentiation also introduces several problems.  
The numerical derivative can amplify noise and requires 
smoothing beforehand, with the final results being highly 
dependent on the parameters of the smoothing algorithm. 
Savitzky-Golay (SG) filtering, based on local least-squares 
fitting of the data by polynomials, is perhaps the most 
well-known method in chemometrics for smoothing and 
computing derivatives on noisy data \cite{sg}.  
Although SG is a common technique for baseline removal, 
SG filtering can unnecessarily reduce the 
signal-to-noise ratio, and is prone to artifacts at the 
end of the wavelength range \cite{schmid}.
Hence, derivative-based baseline removal often amounts  
to a balancing act---it must be smooth enough to 
``clean up'' unwanted noise, but not so much as to remove 
important spectral gradients.


Our particular interest is in the class of derivative smoothers 
that has its roots in the penalized least squares approach of Eilers 
\cite{eilers}. Later penalized variants extended the Eilers 
approach by using weighted least squares generalizations 
that iteratively updated the baseline for a given 
spectrum \cite{eilersals,airpls,arpls}.  
However, what is peculiar about these state-of-the-art penalized baseline 
correction methods is the following observation: they do not 
consider analyte concentrations across samples. 
This is curious because strongly absorbing or scattering 
analytes, possibly distinct from the response variable or 
analyte of interest, can dominate or strongly influence the observed 
spectral variability.\footnote{An earlier 
paper \cite{schecter} did consider analyte concentrations via a 
different class of smoothing, but its regime of applicability was 
quite restrictive: a mixture of solvents in which the concentrations 
of all component species---other than the analyte of interest---is 
known.}
For example, biological samples contain 
considerable moisture content, and water absorbance often dominates 
the observed spectral variability across multiple bands in the NIR 
spectra. However, this moisture information is not considered for 
baseline correction purposes even when reference measurements for
moisture are available.  In short, current baseline correction 
methods are unsupervised in that they are agnostic with respect 
to analyte concentrations.


We propose how current penalized baseline correction methods can 
be modified to accommodate reference measurements associated with 
strongly absorbing or strongly scattering analytes.  We call our 
proposed approach \emph{Supervised Penalized Baseline Correction} 
(SPBC).  In Section \ref{sec:rbc}, we discuss current methods of 
penalized baseline correction.  In Section \ref{sec:SPBC}, we 
propose a modification that can accommodate reference measurements.  
Section \ref{sec:exp} describes the data sets and the performance 
metrics used for assessment, and details the procedure for 
selecting tuning parameters. Section \ref{sec:perf} evaluates 
performance on a suite of baseline correction tasks using two 
NIR data sets.  Section \ref{sec:concl} states the the conclusion 
and suggestions for future work. 


\paragraph{Notation.}
In this paper, matrices and vectors are denoted by uppercase and 
lowercase letters in boldface (e.g., $\bd{X}$ and $\bd{x}$). The 
superscripts $^{\tr}$ and $^+$ indicate the transpose and 
pseudoinverse, respectively, and the superscript $^{-1}$ 
indicates the inverse of a matrix.  The column and row of a 
matrix are denoted by the following subscripts: $\bd{x}_{i:}$ 
is the $i$th row of $\bd{X}$, while $\bd{x}_{:j}$ is the $j$th 
column of $\bd{X}$.  All vectors are column vectors unless 
indicated otherwise.  The comma and semicolon will be used to 
indicate horizontal and vertical concatenation.  For example, 
an $m \times n$ matrix of spectra $\bd{X}$ can be written as $m$ 
samples aligned horizontally 
($\bd{X} = [\, \bd{x}_{1:} \,;\, \ldots \,;\, \bd{x}_{m:}\,]$) 
or as $m$ samples aligned vertically and then transposed 
($\bd{X} = 
[\, \bd{x}_{:1} \,,\, \ldots \,,\, \bd{x}_{:m} \,]^{\tr}$).
Here, each sample is $n$-dimensional: 
$\bd{x}_{:i} = [\, x_{1i} \,;\, \ldots \,;\, x_{ni} \,]$ 
or 
$\bd{x}_{:i} = [\ x_{1i} \,,\, \ldots \,,\, x_{ni} \,]^{\tr}$.
The vector 
$\bd{y} = [\, y_{1} \,;\, \ldots \,;\, y_{m} \,]$
or
$\bd{y} = [\ y_{1} \,,\, \ldots \,,\, y_{m} \,]^{\tr}$
corresponds to an $m \times 1$ vector of reference measurements 
such that $y_i$ corresponds to $\bd{x}_{:i}$.  We will also 
denote another analyte 
$\bd{a} = [\, a_{1} \,;\, \ldots \,;\, a_{m} \,]$
or
$\bd{a} = [\ a_{1} \,,\, \ldots \,,\, a_{m} \,]^{\tr}$
distinct from $\bd{y}$ to indicate a strongly absorbing or 
scattering analyte.

%
\section{Penalized Baseline Correction}
\label{sec:rbc}
%
The approach discussed here relies on penalized least squares 
(or Tikhonov regularization in mathematical parlance) and borrows 
heavily from the algorithmic machinery in \cite{eilers}.  We will 
use the phrase \emph{Penalized Baseline Correction} (PBC) to 
collectively refer to the spectroscopic baseline correction 
approaches discussed by Paul Eilers in \cite{eilers} and later variants 
discussed in Section \ref{sec:we}.

\subsection{Single Spectrum Formulation of Eilers}
\label{sec:ssf}
Suppose $\bd{x}$ indicates a spectrum from a sample and $\bd{z}$ 
denotes the baseline correction vector to be fitted or solved for.
The misfit between $\bd{x}$ and $\bd{z}$ can be expressed as
$\norm{ \bd{x} - \bd{z} }^2$.  However, we want $\bd{z}$ to be 
smooth, and as a result, the roughness can be controlled by 
introducing a penalty term such that we seek to minimize the 
following function \cite{eilers}
\begin{equation}
    f(\bd{z}) = 
    \norm{ \bd{x} - \bd{z} }^2 + 
    \lambda^2 \norm{\bd{D} \bd{z} }^2  
    \,\, = \,\, 
    (\bd{x} - \bd{z})^{\tr}(\bd{x} - \bd{z}) + 
    \lambda^2 
    \bd{z}^{\tr} \bd{C} \bd{z}, 
    \quad
    \bd{C} = \bd{D}^{\tr} \bd{D},
    \,\,
    \lambda > 0,
    \label{eq:Eilers}            
\end{equation}
where the matrix $\bd{D}$ is termed the discrete smoothing operator 
\cite{hansen}.  The matrix $\bd{D}$ typically takes on one of two 
forms---$\bd{D}_1$ or $\bd{D}_2$---where the matrices 
\begin{equation}
    \bd{D}_1 = \matvec{rrrr}{ 1 & -1     &        &  \\
                              & \ddots & \ddots &  \\
                              &        & 1      & -1 }
    \in \mathbb{R}^{(n-1) \times n}
    \quadand
    \bd{D}_2 = \matvec{rrrrr}{ 1 & -2    & 1      &        & \\
                              & \ddots & \ddots & \ddots & \\
                              &        & 1      & -2     & 1 }
    \in \mathbb{R}^{(n-2) \times n}                          
\end{equation}
are scaled approximations to the first and second derivative 
operators.  In the case of the first derivative operator 
where $\bd{D} = \bd{D}_1$, one can express the 
two-norm penalty in Eq.(\ref{eq:Eilers}) as
$\norm{ \bd{D}_1\bd{z} }^2 = \sum_{i=1}^{n-1} (z_i - z_{i+1})^2$.
By setting the gradient of $f(\bd{z})$ in Eq.(\ref{eq:Eilers})
equal to $\bd{0}$,
we arrive at the linear system:
\begin{equation}
    \nabla f(\bd{z}) = \bd{0}
    \quad \Rightarrow \quad
    ( \bd{I} + \lambda^2 \bd{C})\bd{z} = \bd{x}. 
    \label{eq:EilersNormal}
\end{equation}
When $\lambda=0$, then $\bd{z} = \bd{x}$; but this would be a 
non-sensical choice since the baseline-corrected spectra
would be $\bd{x} - \bd{z} = \bd{0}$.  Hence, small values of 
$\lambda$ (i.e., $\lambda << 1$) are not recommended.

\subsection{Weighted Variants}
\label{sec:we}
To introduce flexibility, one can weight the misfit term 
$\norm{ \bd{x} - \bd{z} }^2$ in Eq.(\ref{eq:Eilers})
with a diagonal matrix 
$\bd{H} = \text{diag}(h_1,h_2,\ldots,h_n)$ 
containing non-negative weight entries:
\begin{equation}
    \begin{array}{c}
    {\ds g(\bd{z}) } = 
    \norm{ \bd{H}^{1/2} (\bd{x} - \bd{z}) }^2 + 
    \lambda^2 \norm{\bd{D} \bd{z} }^2      
    \,\, = \,\,
    (\bd{x} - \bd{z})^{\tr} \bd{H} (\bd{x} - \bd{z}) +
    \lambda^2 \bd{z}^{\tr}\bd{C}\bd{z}            
    \\[8pt]
    \nabla {\ds g(\bd{z}) } = \bd{0}
    \quad \Rightarrow \quad
    ( \bd{H} + \lambda^2 \bd{C} ) \bd{z} = \bd{H} \bd{x}.
    \end{array}
    \label{eq:EilersWeighted}
\end{equation}
Subsequent PBC variants of \cite{eilersals,airpls,arpls} 
(known as \ASLS, \AIRPLS\, and \ARPLS, respectively) go much 
further and construct a separate weight matrix for each 
sample $\bd{x}_{:i}$.  Moreover, each sample-specific weight 
matrix is also iteratively updated such that the normal 
equations in Eq.(\ref{eq:EilersWeighted}) become   
\begin{figure}[hb]
\centering
\includegraphics[width = 6.4in]{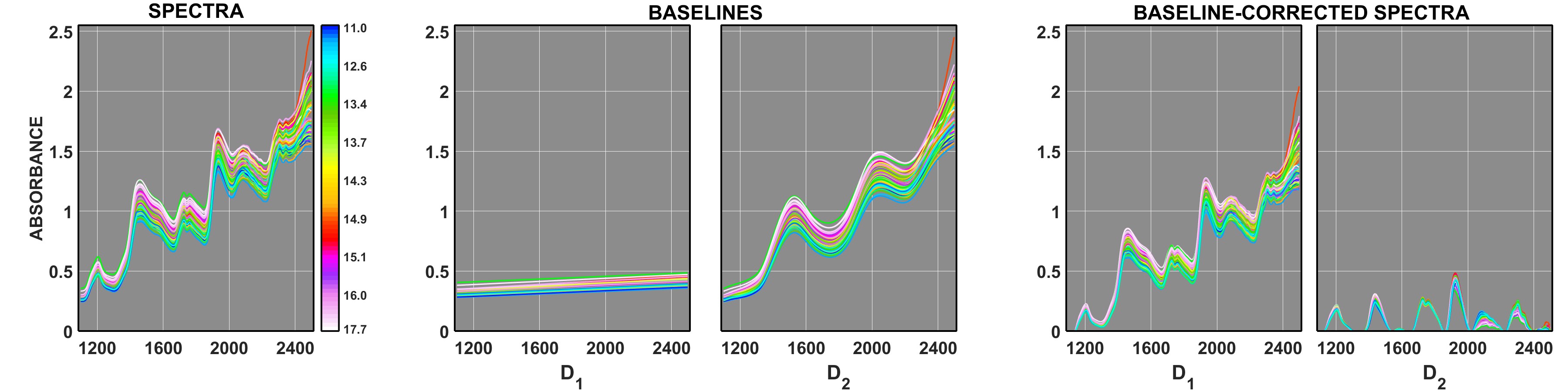}
\caption{\footnotesize For the cookie data set, we
display the spectra (left subplot), 
\AIRPLS\, baselines with $\lambda=100$ via $\bd{D}_1$ 
and $\bd{D}_2$ (middle subplots), and the corresponding 
baseline-corrected spectra (right subplots).}  
\label{fig:CookieBaselineSeq}
\end{figure}
\[
    ( \bd{H}_i^k + \lambda^2 \bd{C} ) \bd{z}_{:i}^k = 
    \bd{H}_i^k \bd{x}_{:i}.
\]
where $i$ and $k$ correspond to the 
$i^{\text{\tiny th}}$ sample and $k^{\text{\tiny th}}$ 
iteration, respectively. Likewise, the baseline vector 
$\bd{z}_{:i}^k = [z_1^k,z_2^k,\ldots,z_n^k]^{\tr}$ 
denotes the baseline-corrected spectrum 
constructed for the $i^{\text{\tiny th}}$ sample
$\bd{x}_{:i} = [x_1,x_2,\ldots,x_n]^{\tr}$ at 
the $k^{\text{\tiny th}}$ iteration.  The $n \times n$ 
diagonal weight matrix is expressed as 
$\bd{H}_i^k =
\text{diag}(h_{i1}^k,h_{i2}^k,\ldots,h_{in}^k)$.
For example, \AIRPLS\, updates the $j^{\text{\tiny th}}$ 
diagonal weight (associated with the $j^{\text{\tiny th}}$ 
wavelength) in the following fashion:
\[
    h_{ij}^k = 
    \left\{ \begin{array}{rl}
    0, &  x_j \geq z_j^{k-1} \\
    \text{exp}(k(x_j - z_j^{k-1})/\rho), & x_j < z_j^{k-1}
    \quad \text{where} \quad
    \rho = {\ds \sum_{l=1}^n \min(0,x_l - z_l^{k-1}). } 
    \end{array} \right.
\]
\ASLS\, and \ARPLS\, use different mechanisms to update the 
diagonal weight entries in $\bd{H}_i^k$.  


Figure \ref{fig:CookieBaselineSeq} illustrates the sequence of 
baseline correction using \AIRPLS: the original spectra, the 
baselines, and the baseline-corrected spectra on the
cookie data set (see Section
\ref{sec:cookie} for a description of this data set). 
The left-most subplot displays the spectra where the 
colored lines indicate the level of water concentration---as 
displayed in the colorbar to the immediate right.
(With respect to baseline correction, 
water is the analyte of interest to be discussed later in this paper.)
The middle two subplots display the baseline spectra constructed 
from $\bd{D}_1$ and $\bd{D}_2$, and the right-most two subplots
display the baseline-corrected spectra for $\bd{D}_1$ and 
$\bd{D}_2$. 
This figure 
highlights the basic question: for regression purposes, is 
it better to use the original spectra $\bd{X}$ or the 
baseline-corrected spectra 
($\bd{X} - \bd{Z}_1$ or $\bd{X} - \bd{Z}_2$)?
The key observation is the following for the variant PBC 
approaches: whereas the Eilers approach applies the same baseline 
correction procedure to each of the $m$ spectra $\bd{x}_{:i}$ 
(via pre-multiplication by $(\bd{I}+\lambda^2\bd{C})^{-1})$, 
the weighted PBC variants perform $m$ different but simultaneous 
baseline corrections in parallel.

\subsection{Multiple Spectrum Formulation}
\label{sec:msf}
Instead of operating on one spectrum at a time, we 
extend Eq.(\ref{eq:Eilers}) and Eq.(\ref{eq:EilersWeighted})
to accommodate an entire $m \times n$ matrix of spectra $\bd{X}$ 
and an entire $m \times n$ matrix of baselines $\bd{Z}$ where 
$\bd{z}_{i:}$ is the baseline associated with $\bd{x}_{i:}$.  
This can be accomplished using the 
Frobenius norm:
\begin{equation}
    f( \bd{Z} ) 
    \,\, = \,\, 
    \fnorm{\bd{X} - \bd{Z}}^2 
    \,+\, 
    \lambda^2 \fnorm{ \bd{DZ}^{\tr} }^2.  
    \label{eq:EilersMult}              
\end{equation}
The Frobenius norm of an $m \times n$ 
matrix $\bd{A}$ is expressed as 
$\fnorm{\bd{A}}^2 = \sum_{i=1}^m \sum_{j=1}^n a_{ij}^2$
and can be thought of as a two-norm on the ``flattened version'' 
of $\bd{A}$ where the flattened vector $\bd{A}$ now 
has size $mn \times 1$.  Setting the gradient of 
Eq.(\ref{eq:EilersMult}) equal to zero (in addition  
to its weighted equivalent in Eq.(\ref{eq:EilersWeighted})), 
we obtain the subsequent normal equations 
\cite{cookbook}: 
\begin{equation}  
    \begin{array}{rcc}
    \nabla f(\bd{Z}) = \bd{0}
    & \quad \Rightarrow \quad & 
    \bd{Z}(\bd{I} + \lambda^2 \bd{C}) = \bd{X}, 
    \\[8pt]
    \nabla {\ds g(\bd{Z}) } = \bd{0}
    & \quad \Rightarrow \quad & 
    \bd{Z}(\bd{H} + \lambda^2 \bd{C}) = \bd{XH}. 
    \end{array}
    \label{eq:EilersMultNormal}
\end{equation}
The equations in Eq.(\ref{eq:EilersMultNormal}) 
are essentially the same as in 
Eqs.(\ref{eq:EilersNormal},\ref{eq:EilersWeighted})
but the coefficient matrices
$\bd{I} + \lambda^2 \bd{C}$ and 
$\bd{H} + \lambda^2 \bd{C}$ are 
applied to all baseline spectra simultaneously as opposed to one 
spectrum at a time.  Note that in 
Eqs.(\ref{eq:EilersNormal},\ref{eq:EilersWeighted}), the 
spectra $\bd{x}$ and $\bd{z}$ are column vectors while the 
collective spectra in $\bd{X}$ and $\bd{Z}$ are aligned 
row-wise.  To maintain alignment consistency with 
Eqs.(\ref{eq:EilersNormal},\ref{eq:EilersWeighted}), one could 
rewrite the equations in a column-wise format, e.g., 
$(\bd{I} + \lambda^2 \bd{C}) \bd{Z}^{\tr}= \bd{X}^{\tr}$
and 
$(\bd{H} + \lambda^2 \bd{C})\bd{Z}^{\tr} = \bd{HX}^{\tr}$.

%
\section{Supervised Penalized Baseline Correction} 
\label{sec:SPBC}
%
In Sections \ref{sec:ssf} and \ref{sec:msf}, only the 
matrix $\bd{X}$ is used to construct the baseline matrix
$\bd{Z}$.  However, the approach in Section \ref{sec:msf}
can be modified to accommodate a priori analyte 
information.  The forthcoming \emph{supervised PBC} approaches will
be denoted by the acronym SPBC.  The first SPBC approach 
is based on Nonlinear Iterative Partial Least Squares (NIPALS) 
and will be denoted as \SPBCN.
The second approach is based on Inverse Least Squares (ILS) 
and will be denoted as 
\SPBCI. 

%
\subsection{NIPALS framework of \SPBCN}
%
Let the vector $\bd{a} = [a_1,a_2,\ldots,a_m]^{\tr}$ denote an analyte 
that will be used to construct the baseline. Here, we extend the Eilers 
approached via the NIPALS outer-product approach
\begin{equation}
    \begin{array}{rclcl}
    f(\bd{w},\bd{Z}) 
    & = &  
    \fnorm{ (\bd{X} - \bd{Z})  - \bd{aw}^{\tr} }^2 
    & + &  
    \lambda^2 \fnorm{ \bd{DZ}^{\tr} }^2 
    \\[8pt]
    & = & 
    \fnorm{ \bd{R}  - \bd{Z} }^2 
    & + & \lambda^2 \fnorm{ \bd{DZ}^{\tr} }^2,
    \,\, 
    \bd{R} = \bd{X} - \bd{a} \bd{w}^{\tr}.
    \end{array}
    \label{eq:SPBCn}
\end{equation}
Note that the Eilers approach of Eq.(\ref{eq:EilersMult}) and the NIPALS 
extension in Eq.(\ref{eq:SPBCn}) are functionally equivalent with $\bd{X}$ 
being swapped out for with $\bd{R}$ in Eq.(\ref{eq:SPBCn}).  In effect, 
\SPBCN\, baseline-corrects the \emph{residual} or deflated matrix $\bd{R}$ 
instead of $\bd{X}$.  (When $\bd{a} = \bd{0}$, \SPBCN\, reduces to the 
Eilers approach.)  Since Eq.(\ref{eq:SPBCn}) is now a function of two 
variables $\bd{Z}$, we set the gradients of $f(\bd{w},\bd{Z})$---separately 
with respect to $\bd{w}$ and $\bd{Z}$---equal to zero and obtain:
\begin{equation}
    \begin{array}{rcrcl}
    {\ds \nabla_{\bd{w}} } f = 0 
    & \quad \Rightarrow \quad & 
    \bd{w} 
    & = &   
    {\ds 
    \frac{ (\bd{X}  - \bd{Z})^{\tr} \bd{a} }
         { \bd{a}^{\tr} \bd{a} }
    }
    \\[8pt]
    {\ds \nabla_{\bd{Z}} } f = 0 
    & \quad \Rightarrow \quad & 
    \bd{Z} 
    & =  &
    \bd{R} (\bd{I} + \lambda^2 \bd{C})^{-1}.
    \end{array}
    \label{eq:pseudo-N}
\end{equation}
The above equations can now be solved via alternating least squares (ALS): 
solve for $\bd{w}$ in the ${\ds \nabla_{\bd{w}} } f = 0$ step, plug in
the resultant $\bd{w}$ in the equations associated with 
${\ds \nabla_{\bd{Z}} } f = 0$ and solve for $\bd{Z}$.
The pseudocode for this ALS approach is given in 
Algorithm \ref{alg:SPBCn}.  The most computationally intensive 
step in the pseudocode occurs in Step 4, i.e., 
solve 
$\bd{Z}_{(k+1)} = \bd{R} (\bd{I} + \lambda^2 \bd{C})^{-1}$.
In the classical PBC approach of Eilers in Eq.({\ref{eq:Eilers}}),
sparse matrix linear libraries coupled with 
Cholesky factorization was used to efficiently 
solve the linear system.  
However, a much faster numerical implementation can be performed, 
particularly in the case of $\bd{D} = \bd{D}_1$; see Section
\ref{sec:ffspbc} of the Supplement.
\begin{algorithm}
    \caption{\quad \SPBCN}
    \label{alg:SPBCn}
    \begin{algorithmic}
    \State {\small 
            \textbf{input}: 
            $\bd{X}$, 
            $\bd{a}$, 
            $\lambda$, 
            $\bd{C} = \bd{D}^{\tr} \bd{D}$
            }
    \State {\small 
            \textbf{initialize}: 
            $k=0$, 
            $\bd{Z}_{(k)} = \bd{0}$
            }
    \While{not converged}
        \State {\small 
                \textbf{Step 1}: 
                $\bd{B} = \bd{X} - \bd{Z}_{(k)}$
                }
            \State {\small 
                \textbf{Step 2}: 
                $\bd{w} = 
                {\ds \frac{ \bd{B}^{\tr} \bd{a} } 
                     { \bd{a}^{\tr} \bd{a}} }$
                }                  
        \State {\small 
                \textbf{Step 3}: 
                $\bd{R} = \bd{X} - \bd{aw}^{\tr}$
                }            
        \State {\small 
                \textbf{Step 4}: 
                $\bd{Z}_{(k+1)} = 
                 \bd{R} (\bd{I} + \lambda^2 \bd{C})^{-1}$
                }
        \State {\small 
                \textbf{Step 5}: 
                $k = k+1$
                }
    \EndWhile
    \State \textbf{return} $\bd{Z} = \bd{Z}_{(k+1)}$
    \end{algorithmic}
\end{algorithm}

\begin{figure}[ht]
    \centering
    \includegraphics[width=6.4in]{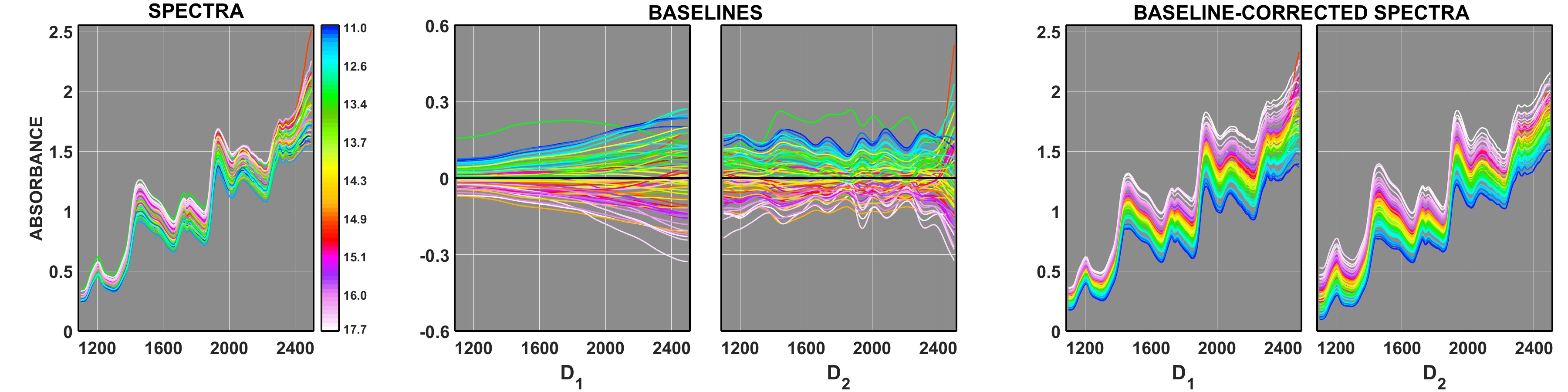}
    \caption{\footnotesize 
    Spectra, baseline spectra and baseline-corrected spectra for 
    the first and second derivative operators 
    ($\bd{D}_1$ and $\bd{D}_2$) when using \SPBCN.}
\label{fig:cookie_spbc}
\end{figure}

Figure \ref{fig:cookie_spbc} gives an example of the \SPBCN-based 
baseline correction process for the cookie data set 
where water concentrations in $\bd{a}$ are 
used to construct the baselines.  Compared to 
Figure \ref{fig:CookieBaselineSeq} where baseline correction 
via \AIRPLS\, was performed, the 
baseline-corrected spectra via SPBC exhibit a more sequential 
arrangement of spectra as a function of water concentration---as
absorbance increases for a particular wavelength, the spectral 
values increase in concentration.

\subsection{ILS framework of \SPBCI}
Instead of the outer product approach of \SPBCN, we can employ an 
inner product approach via ILS to extend Eq.(\ref{eq:EilersMult}):
\begin{equation}
    \begin{array}{rclcl}
    f(\bd{w},\bd{Z}) 
    & = & 
    \norm{ (\bd{X} - \bd{Z})\bd{w}  - \bd{a} }^2 
    & + &  
    \lambda^2 \fnorm{ \bd{DZ}^{\tr} }^2 
    \\[8pt]
    & = &  
    \norm{ \bd{Zw} - \bd{r} }^2 
    & + &  
    \lambda^2 \fnorm{ \bd{DZ}^{\tr} }^2, \quad
    \bd{r} = \bd{Xw} - \bd{a}.
    \end{array}
    \label{eq:SPBCi}
\end{equation}
Here, we are trying to relate the baseline corrected spectra
$\bd{X} - \bd{Z}$ to the analyte concentrations in $\bd{a}$
via the regression vector $\bd{w} = [w_1,w_2,\ldots,w_n]^{\tr}$.
As with Eq.(\ref{eq:SPBCn}), we set of the gradients,
separately with respect to $\bd{w}$ and $\bd{Z}$,
equal to zero and obtain:
\begin{equation}
    \begin{array}{rcccl}
    {\ds \nabla_{\bd{w}} } f = 0 
    & \quad \Rightarrow \quad & 
    \bd{w}
    & = &     
    \left[(\bd{X}  - \bd{Z})^{\tr} (\bd{X}  - \bd{Z}) \right]^{+}
    (\bd{X}  - \bd{Z})^{\tr} \bd{a}
    \\ [3pt]
    {\ds \nabla_{\bd{Z}} } f = 0 
    & \quad \Rightarrow \quad & 
    \bd{Z}
    & =  &
    \bd{rw}^{\tr} (\bd{ww}^{\tr} + \lambda^2 \bd{C})^{-1}.
    \end{array}
\end{equation}
The pseudocode for \SPBCI\, via ALS is given in Algorithm 2.
The most computationally intensive 
steps in the pseudocode occurs in Steps 2 and 4, i.e., solve 
$\bd{B}^{\tr} \bd{B} \bd{w} = \bd{B}^{\tr} \bd{a}$ for $\bd{w}$
and 
$\bd{Z}_{(k+1)} = 
\bd{rw}^{\tr} 
\left( \bd{ww}^{\tr} + \lambda^2 \bd{C} \right)^{-1}$,
respectively.  See Section \ref{sec:ffspbc} in the Supplement for 
details on how these steps were numerically implemented.
\begin{algorithm}
    \caption{\quad \SPBCI}
    \label{alg:SPBCi}
    \begin{algorithmic}
    \State {\small 
            \textbf{input}: 
            $\bd{X}$, 
            $\bd{a}$, 
            $\lambda$, 
            $\bd{C} = \bd{D}^{\tr} \bd{D}$
            }
    \State {\small 
            \textbf{initialize}: 
            $k=0$, 
            $\bd{Z}_{(k)} = \bd{0}$, 
            $\bd{w}_{(k)} = \bd{0}$
            }
    \While{not converged}
        \State {\small 
                \textbf{Step 1}: 
                $\bd{B} = \bd{X} - \bd{Z}_{(k)}$
                }
        \State {\small 
                \textbf{Step 2}: 
                Solve $\bd{B}^{\tr} \bd{B} \bd{w} = 
                \bd{B}^{\tr} \bd{a}$
                for $\bd{w}$
                }
        \State {\small 
                \textbf{Step 3}: 
                $\bd{r} = \bd{Xw} - \bd{a}$
                }            
        \State {\small 
                \textbf{Step 4}: 
                $\bd{Z}_{(k+1)} = 
                \bd{rw}^{\tr} 
                \left( \bd{ww}^{\tr} + 
                \lambda^2 \bd{C} \right)^{-1}$
                }
        \State {\small 
                \textbf{Step 5}: 
                $k = k+1$
                }
    \EndWhile
    \State \textbf{return} $\bd{Z} = \bd{Z}_{(k+1)}$
    \end{algorithmic}
\end{algorithm}

\subsection{Sample Dependence}
In the Eilers approach where 
$\bd{z}_{:i} = 
(\bd{I} + \lambda^2 \bd{C})^{-1} \bd{x}_{:i}^{\tr}$,
the baseline correction procedure for each spectrum 
$\bd{x}_{:i}^{\tr}$ is the same, i.e., pre-multiplication by 
$(\bd{I} + \lambda^2 \bd{C})^{-1}$.
In the weighted variants (e.g., \ASLS\, \AIRPLS\, or \ARPLS),  
$\bd{z}_{:i} = 
(\bd{H}_i^k + \lambda^2 \bd{C})^{-1} \bd{H}_i^k \bd{x}_{:i}^{\tr}$,
and as a result, the baseline correction procedure is the not the 
same for each spectrum.  However, like the Eilers approach, 
baseline correction for any one spectrum $\bd{x}_{:i}$ can be done 
in parallel, (i.e., the baseline correction done one spectrum 
does not depend on the baseline correction done an another 
spectrum). Baseline correction for SPBC approaches, on the other 
hand, cannot be done one spectrum at a time. They must be done 
in batch fashion. Thus, the baseline for each sample encodes 
information (on analyte concentration) across the entire 
calibration set which is in contrast to previous approaches.

%
\section{Experimental Methods} 
\label{sec:exp}
%

\subsection{Data Sets}
\begin{figure}[ht]
    \centering
    \includegraphics[width = 6.5in]{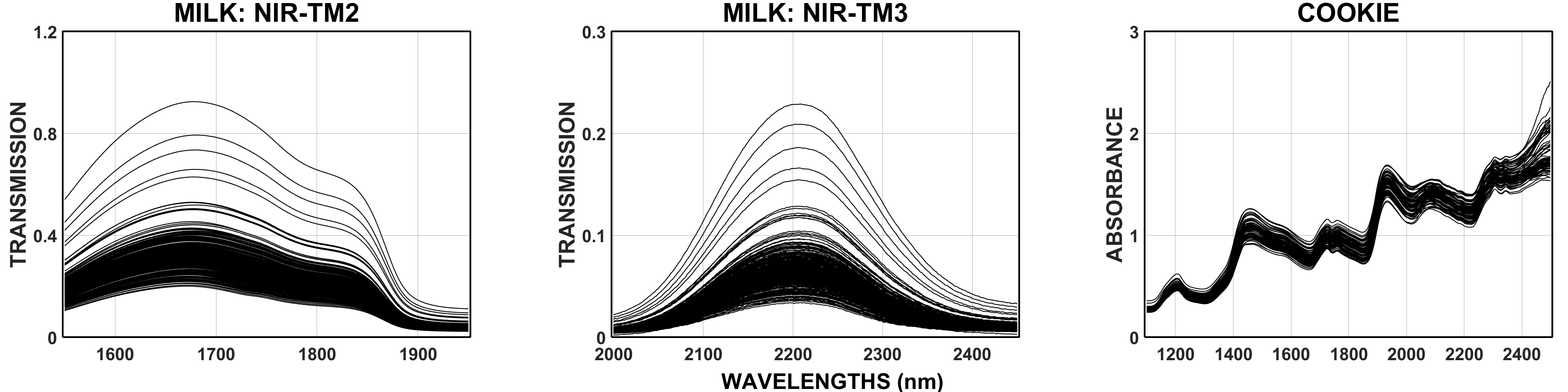}
    \caption{\footnotesize Spectra for the milk data set 
    (instruments NIR-TM2 and NIR-TM3 in transmission mode) and 
    the cookie data set (in absorbance mode) on the far right.}  
    \label{fig:Spectra}
\end{figure}

We will examine two near infrared (NIR) data sets: 
the milk and cookie data 
sets.  The NIR spectra for these data sets are displayed in Figure 
\ref{fig:Spectra}. 

\subsubsection{Cookie Data Set}
\label{sec:cookie}
The cookie data set contains measurements from quantitative NIR 
spectroscopy \cite{cookie}.  The intent of using this data set is to 
test the feasibility of NIR spectroscopy to measure the 
composition of biscuit dough pieces. There are four analytes 
under consideration: fat, sucrose, flour, and water. The 
calculated percentages of these four ingredients represent the 
four response variables. There are 72 samples in total: 40 samples in 
the calibration set (with sample 23 being an outlier) and 32 
samples in the separate prediction or validation set (with 
example 21 considered as an outlier).  An NIR reflectance 
spectrum is available for each dough piece. The spectral data 
consist of 700 points measured from 1100 to 2498 nanometers (nm) 
in intervals of 2nm.  In this data set, sucrose will be the 
response variable ($\bd{y}$) to be predicted, while fat, water 
and flour will each separately be the analyte $\bd{a}$ that will 
be used to construct the baselines.  

\subsubsection{Milk Data Set}
The milk data set consists of 298 samples measured across three 
separate Microelectromechanical System (MEMS) NIR spectrometers 
in transmission mode \cite{milk}.  The three spectrometers are 
denoted in this paper as NIR-TM1, NIR-TM2 and NIR-TM3. The 
spectrum for each milk sample is an average of 20 replicates.
NIR-TM1, NIR-TM2 and NIR-TM3 span 1100-1400nm, 1550-1950nm and 
2000-2450nm, respectively, with an interval of 2nm.  There are 
six primary analytes under consideration: fat, lactose, protein,
urea, solute and dry matter. 
We will focus on instruments NIR-TM2 and NIR-TM3.  
In this data set, fat will be the analyte ($\bd{a}$) that will 
be used to construct the baselines.  Lactose, protein, urea, 
solute and dry matter will each separately be the response 
variable or analyte $\bd{y}$ to be predicted.

\subsection{Schemes involving the availability of $\bd{a}$}
\label{sec:schemes}
The SPBC implementation depends on how much
information associated with the analyte $\bd{a}$ is 
available.  Data-wise, we will use the triplet 
$\{\bd{X},\bd{a},\bd{y}\}$. 
The $m \times n$ matrix $\bd{X}$ denotes the spectra to be 
baseline corrected, the $m \times 1$ vector $\bd{a}$ 
corresponds to the analyte that will be used for baseline 
correction, and the $m \times 1$ vector $\bd{y}$ corresponds 
to the response variable or analyte whose concentrations we 
want to predict.  We will split the data into three parts: 
the calibration (or training), tuning, and validation (or 
test) sets, which will be denoted by the subscripts 
\texttt{1}, \texttt{t} and \texttt{2}, i.e., 
$\{ \bd{X}_1,\bd{a}_1,\bd{y}_1 \}$,
$\{ \Xtune,\atune,\ytune \}$ and 
$\{ \bd{X}_2,\bd{a}_2,\bd{y}_2 \}$.
The tuning set will be aside and will be exclusively used to 
estimate the number of PLS latent dimensions. See Section 
\ref{sec:dp} for a more detailed explanation of how the data 
is partitioned.  Ultimately, our goal is to enhance the 
prediction of $\bd{y}_2$ by utilizing baseline corrected 
spectra constructed from
$\bd{X} := [\bd{X}_1; \bd{X}_2]$ and 
$\bd{a} := [\bd{a}_1; \bd{a}_2]$. 
(The symbol ``:='' typically denotes that the left-hand side 
is defined as the expression on the right-hand side.) The 
prediction of $\bd{y}_2$ proceeds in two steps to be 
described next.


\subsubsection{Full and Partial Schemes} 
We use 
$\bd{X} := [\bd{X}_1; \bd{X}_2]$ and 
$\bd{a} := [\bd{a}_1; \bd{a}_2]$
in algorithms \ref{alg:SPBCn} or \ref{alg:SPBCi} to 
obtain $\bd{Z}$, and then split it into two parts 
$\bd{Z}_1$ and $\bd{Z}_2$
corresponding to the calibration and validation sets
such that $\bd{Z} := [\bd{Z}_1;\bd{Z}_2]$.  Computing 
$\bd{Z}_1$ and $\bd{Z}_2$ requires $\bd{a}_1$ and $\bd{a}_2$, 
respectively. 
The \emph{full scheme} assumes
that we have full access to both $[\bd{X}_1; \bd{X}_2]$
and $[\bd{a}_1; \bd{a}_2]$.
(Suppose the reference measurements for both $\bd{a}_1$ 
and $\bd{a}_2$ are inexpensive and/or easy to obtain with respect 
to laboratory effort and time, then 
$\bd{X} := [\bd{X}_1;\bd{X}_2]$ and 
$\bd{a} := [\bd{a}_1;\bd{a}_2]$ 
will be the inputs into Algorithms 1 and 2).  The 
\emph{partial scheme} assumes that we have full access to 
$\bd{X} := [\bd{X}_1; \bd{X}_2]$ but only partial access to 
$\bd{a}$, that is we have knowledge of $\bd{a}_1$ but 
not $\bd{a}_2$.
Without access to $\bd{a}_2$, however, we will need reliable 
approximations or estimates to act as numerical proxies. 
Instead of using 
$\bd{a} := [\bd{a}_1;\bd{a}_2]$, we can use a combined set 
of known references $\bd{a}_1$ and prediction estimates 
$\estav$ such that $\widehat{\bd{a}} := [\bd{a}_1;\estav]$.  
In short, for the partial scheme, we use $\widehat{\bd{a}}$ 
to construct the baselines instead of $\bd{a}$.  
Compared to the partial scheme, we can expect the construction 
of the baseline spectra for the full scheme to be qualitatively 
better since known references are used.
Hence, the performance of the partial scheme will be highly 
dependent on the accuracy and precision associated with the 
estimates in $\estav$.


The construction of the estimates in $\estav$ proceeds as follows
for each data partition.  80\% of the samples are randomly sampled 
from the calibration set $\{\bd{X}_1,\bd{a}_1\}$.  The calibration 
model is then applied to $\bd{X}_2$ and a prediction estimate 
$\widehat{a}_2^{(1)}$ is obtained.  Another 80\% of the samples 
are randomly sampled from the calibration set, the 
subsequent model is then applied to $\bd{X}_2$ and another prediction 
estimate $\widehat{a}_2^{(2)}$ is obtained.   This process is 
repeated for a total of 25 times such that we obtain the following 
collection of estimates 
$\{ \widehat{a}_2^{(1 )}, \widehat{a}_2^{(2)},\ldots,
    \widehat{a}_2^{(25)} \}$.
The prediction estimates outside the ``Tukey interval'' (or
$[Q_1 - 1.5(Q_3-Q_1),Q_3 + 1.5(Q_3-Q_1)]$) are removed and the 
remaining estimates are averaged to yield the final estimate for
$\estav$.

          
\subsubsection{Build calibration model and predict $\bd{y}_2$}
Once $\bd{Z} = [\bd{Z}_1;\bd{Z}_2]$ has been obtained, we 
baseline-correct the calibration and validation sets
whereby
$\Xcorr{1} = \bd{X}_1 - \bd{Z}_1$
and 
$\Xcorr{2} = \bd{X}_2 - \bd{Z}_2$, 
respectively. We mean-center the calibration set
\[ \begin{array}{rl}
    \bd{\mu}_x = ( \bd{1}_{n_1}^{\tr}\Xcorr{1} )/n_1, \quad & 
    \Xcorr{1} := \Xcorr{1} - \bd{1}_{n_1} \bd{\mu}_x
    \\[3pt]
    \bd{\mu}_y = (\bd{1}_{n_1}^{\tr}\bd{y}_1)/n_1, \quad & 
    \bd{y}_1 := \bd{y}_1 - \bd{1}_{n_1} \bd{\mu}_y,
\end{array} \]
and solve $\Xcorr{1} \bd{b} = \bd{y}_1$ for 
$\bd{b}$ using, for example, Partial Least Squares (PLS) regression. 
Finally, we then predict $\bd{y}_2$ via 
\[
    \hat{\bd{y}}_2 = 
    ( \Xcorr{2} - \bd{1}_{n_2}\bd{\mu}_x ) \bd{b} + 
    \bd{1}_{n_2} \bd{\mu}_y.
\]

\subsection{Baseline Correction Methods Examined}
We will examine several classes of penalized smoothing methods: 
1) no background correction (just using the original spectra 
without pre-processing); 
2) the original PBC approach of Eilers in Section \ref{sec:ssf}; 
3) a PBC smoothing variant of Section \ref{sec:we}; and
3) the SPBC methods introduced in Section \ref{sec:SPBC}.  
We outline them below:
\begin{itemize}
    \item 
    \NONE: 
    Here, no background correction is applied.  However, from 
    a background correction point of view, $\bd{Z} = \bd{0}$
    and the baseline corrected spectra is simply 
    $\bd{X} - \bd{Z} = \bd{X}$.  
    \NONE\, then serves as the benchmark by which the other 
    baseline correction methods are intended to outperform.
    \item
    \EILERS: 
    This refers to the construction of the baseline spectra 
    $\bd{Z}$ by the original PBC approach of Eilers in Section 
    \ref{sec:ssf}.
    \item 
    \AIRPLS: 
    The baseline spectra $\bd{Z}$ are constructed via Adaptive 
    Iteratively Reweighted Penalized Least Squares \cite{airpls}.  
    With respect to the other PBC variants mentioned in Section 
    \ref{sec:we} (\ASLS\, and \ARPLS) that use weighted least 
    squares, we observed that these variants performed 
    qualitatively the same as \AIRPLS. As a result, and for ease 
    of illustration, we use \AIRPLS\, as the canonical PBC 
    variant. 
    \item 
    \SPBC:
    The SPBC methods construct the baseline spectra $\bd{Z}$ 
    by accommodating analyte information. The SPBC approaches 
    can be subdivided by approach (inverse least squares versus 
    NIPALS) and by scheme (full versus partial):
    \begin{itemize}
        \item[$\circ$] \SPBCIF: 
        Inverse least squares coupled with the full scheme.
        \item[$\circ$] \SPBCIP: 
        Inverse least squares coupled with the partial scheme.
        \item[$\circ$] \SPBCNF: 
        NIPALS coupled with the full scheme.
        \item[$\circ$] \SPBCNP: 
        NIPALS coupled with the partial scheme.
    \end{itemize}
\end{itemize}
We also explored the smoothing approaches of Savitsky-Golay 
(\SG) and Extended Multiplicative Scatter Correction (\EMSC) 
\cite{sg,msc,emsc}.  Here, our version of \EMSC\, utilizes 
a ``plain vanilla'' approach that accounts for wavelength 
dependencies where the fitting coefficients 
$\boldsymbol{\beta}$ were modeled as
\[
\bd{X} \boldsymbol{\beta} = 
[\bd{1}, \bd{r}, \boldsymbol{\lambda}, \boldsymbol{\lambda}^2].
\]
Here, $\bd{r}$ is the reference spectrum and 
$\boldsymbol{\lambda} = 
[\lambda_1,\lambda_2,\ldots,\lambda_n]^{\tr}$
is the vector of wavelengths.  Although we have knowledge of
the concentrations of many analytes, we do not assume that we 
have enough knowledge across the major chemical constituents 
(analytes and interferents) in the milk and cookie data sets; 
hence the rationale for employing the basic EMSC approach accounting 
only for wavelength dependencies.  We found that \SG\, and \EMSC\, 
were inferior to \AIRPLS\, in all instances (and in the case of 
\SG, we even tried to optimize for frame length, or moving 
window width).  As a consequence, and as was the case with 
\ASLS\, and \ARPLS, we also do not display performance results 
for \SG\, and \EMSC.

\subsection{Data Partitions and Assessment Metrics}
\label{sec:dp}
To ensure that performance results are not anecdotal to one 
particular split of the data, we assess the performance across 
200 splits of the data.  Each partition of the $m$ samples 
randomly shuffles the data and splits it into three sets: 
45\% (calibration), 5\% (tuning) and 50\% (validation or 
testing).  The first 45\% of the samples will be used to build 
the calibration model. The next 5\% of the samples belong to 
the tuning set.  The prediction of $\bd{y}$ on the tuning set 
samples will be used to select the PLS latent dimension that 
will subsequently be applied to the validation set. 
Aside from the tuning set, we split the samples into two 
sets of triplets: the 
calibration triplet $\{ \bd{X}_1,\bd{y}_1,\bd{a}_1 \}$---derived 
from the 45\% block of samples---and the validation triplet
$\{ \bd{X}_2,\bd{y}_2,\bd{a}_2 \}$---derived from the 50\% block 
of samples.  Note that the SPBC partial 
scheme uses the validation triplet 
$\{ \bd{X}_2,\bd{y}_2,\estav \}$ 
where $\estav$ is a proxy or prediction estimate for $\bd{a}_2$.


To assess the performance for the $i^{\text{\tiny th}}$ partition 
or data split, we use two metrics: MARD and the coefficient of 
determination ($R^2$).  MARD is an acronym for Mean Absolute 
Relative Difference, and is computed as the mean value of the 
absolute relative difference (ARD) between prediction estimates 
and reference measurements.  For example, MARD for the validation 
set would be computed as follows: the predictions and reference 
measurements for the $i^{\text{\tiny th}}$ partition are defined as  
$\widehat{\bd{y}}_2 = 
[\hat{y}_{2,1}^{(i)},\ldots,\hat{y}_{2,m_2}^{(i)}]^{\tr}$ 
and 
$\bd{y}_2 = [y_{2,1}^{(i)},\ldots,y_{2,m_2}^{(i)}]^{\tr}$, 
respectively, and
\[
    \text{ARD}_j^{(i)} = 
    100\% 
    \left| 
    \frac{ \hat{y}_{2,j}^{(i)} - y_{2,j}^{(i)} }{ y_{2,j}^{(i)} }  
    \right|, 
    \quad\
    \text{MARD}^{(i)} = 
    \frac{1}{m_2} \sum_{j = 1}^{m_2} \text{ARD}_k^{(i)}.
\]
To compute MARD for the tuning set, one would instead replace 
$\widehat{\bd{y}}_2$ and $\bd{y}_2$ with
$\widehat{\bd{y}}_{\mbox{\texttt{\scriptsize t}}} 
= 
[\hat{y}_{\texttt{\scriptsize t},1},\ldots,
 \hat{y}_{\texttt{\scriptsize t},m_t}]^{\tr}$ 
and 
$\ytune = 
[y_{\texttt{\scriptsize t},1},\ldots,
 y_{\texttt{\scriptsize t},m_t}]^{\tr}$,
respectively.  MARD basically functions as an aggregate percent 
relative error measure across a set of samples. The coefficient 
of determination metric derives from the line-of-best-fit in the 
scatter diagram associated with the coordinates
\[ \{ \, (y_{2,1}^{(i)},\hat{y}_{2,1}^{(i)}), \,
       (y_{2,2}^{(i)},\hat{y}_{2,2}^{(i)}), \, 
       \ldots, \,
       (\hat{y}_{2,m_2}^{(i)},\hat{y}_{2,m_2}^{(i)}) 
\, \} \]    
between the reference measurements and prediction estimates. 
The coefficient of determination for the $i^{\text{th}}$ 
partition will be denoted as $\text{R2}^{(i)}$.  We then create 
boxplots from the collection of $\text{MARD}^{(i)}$ and 
$\text{R2}^{(i)}$ measures across the partitions 
$i \in \{1,2,\ldots,200\}$.  Instead of the traditional boxplots
where the inter-quartile range is the middle 50\% of the data, 
we modify our boxplots to show the middle 80\% where the edges 
of the ``box'' correspond to the 10\% and 90\% percentiles.  
Moreover, no outliers are displayed; instead the whiskers extend 
to the min and max of the data.

\subsection{Selection of $\lambda$ values}
\label{sec:lambda}
In the penalized methods associated with Eilers, the PBC 
variants such as \AIRPLS, and the SPBC approaches, the 
value of $\lambda$ is the tuning parameter of interest.  The 
simplicity of the Eilers approach, i.e., 
$\bd{Z} (\bd{I} + \lambda^2 \bd{C}) = \bd{X}$,
yields insight on what a 
reasonable choice $\lambda$ should be.   
When $\lambda$ is small ($0 < \lambda \ll 1$), 
then $\bd{Z} \approx \bd{X}$ and the  
baseline corrected spectra $\bd{X} - \bd{Z}$ will essentially 
be small-amplitude noise around the zero matrix.  Hence, small 
values of $\lambda$ are not warranted.
The solution of 
$\bd{Z} (\bd{I} + \lambda^2 \bd{C}) = \bd{X}$
is equivalent to a sum involving the loading 
vectors $\bd{v}_{:j}$ of the derivative operator $\bd{D}$---see 
Eq.(\ref{eq:ffeilers}) in the Supplement. 
The filter factors $f_j = 1 / (1 + \lambda^2 s_j^2)$ 
in Eq.(\ref{eq:ffeilers}) can only 
damp or filter the corresponding loading vector 
$\bd{v}_{:j}$ when $\lambda$ is sufficiently large, i.e. 
($\lambda \gg 1$).  As result, we will assess performance across 
four penalty values: $\lambda = \{1, 10, 100, 1000\}$. 

\subsection{Selection of the latent dimension}
\label{sec:latent}
As mentioned in Section \ref{sec:schemes}, the calibration model 
required for predicting $\bd{y}_2$ in Step 2 in the full and 
partial schemes in Section \ref{sec:schemes} will be done using 
Partial Least Squares (PLS).  To select the PLS latent dimension, 
we use an approach based on metric ranking.


Based upon the predictions on the tuning set, let's consider the 
MARD and R2 values across PLS latent dimensions $1,2\ldots,20$.
The latent dimension with the lowest MARD value gets a rank of 1; 
the latent dimension with the second lowest MARD value gets a rank 
of 2; and so on.  Similarly, the latent dimension with the highest 
R2 value gets a rank of 1; the latent dimension with the second 
highest R2 value gets a rank of 2; and so on.  Let 
$[\alpha_1,\alpha_2,\ldots,\alpha_{20}]$
and 
$[\beta_1,\beta_2,\ldots,\beta_{20}]$ 
correspond to the integer-based rankings associated with MARD and 
R2, respectively, across PLS latent dimensions $1,2\ldots,20$.  
Hence, each latent dimension $k$ is associated with a pair of 
ranks $(\alpha_k,\beta_k)$, and we can treat this pair as 
$x$- and $y$-coordinates.  The PLS latent dimension $k$ whose 
coordinates $(\alpha_k,\beta_k)$ is closest to the origin 
$(0,0)$---using the Euclidean distance 
$\sqrt{\alpha_k^2 + \beta_k^2}$---is deemed the optimal 
PLS latent dimension.

%
\section{Performance} 
\label{sec:perf}
%

In this section, we examine performance for both the Milk and 
Cookie data sets.  A collection of MARD and R2 values across 
200 data partitions will be used to assess performance.  


\subsection{Milk data set performance}
\label{sec:MilkPerf}


For the Milk data set, fat will be the analyte $\bd{a}$ used 
(in tandem with the spectra $\bd{X}$) to construct the baseline 
spectra $\bd{Z}$.  Prediction will first be assessed on urea.  
Performance will then be examined for all of the 
other analytes in order of their correlation strength with fat.

\subsubsection{Fat ($\bd{a}$) and Urea ($\bd{y}$)}
\label{sec:Milk-Fat-Urea}

\begin{figure}[th]
    \centering
    \includegraphics[width=6.5in]{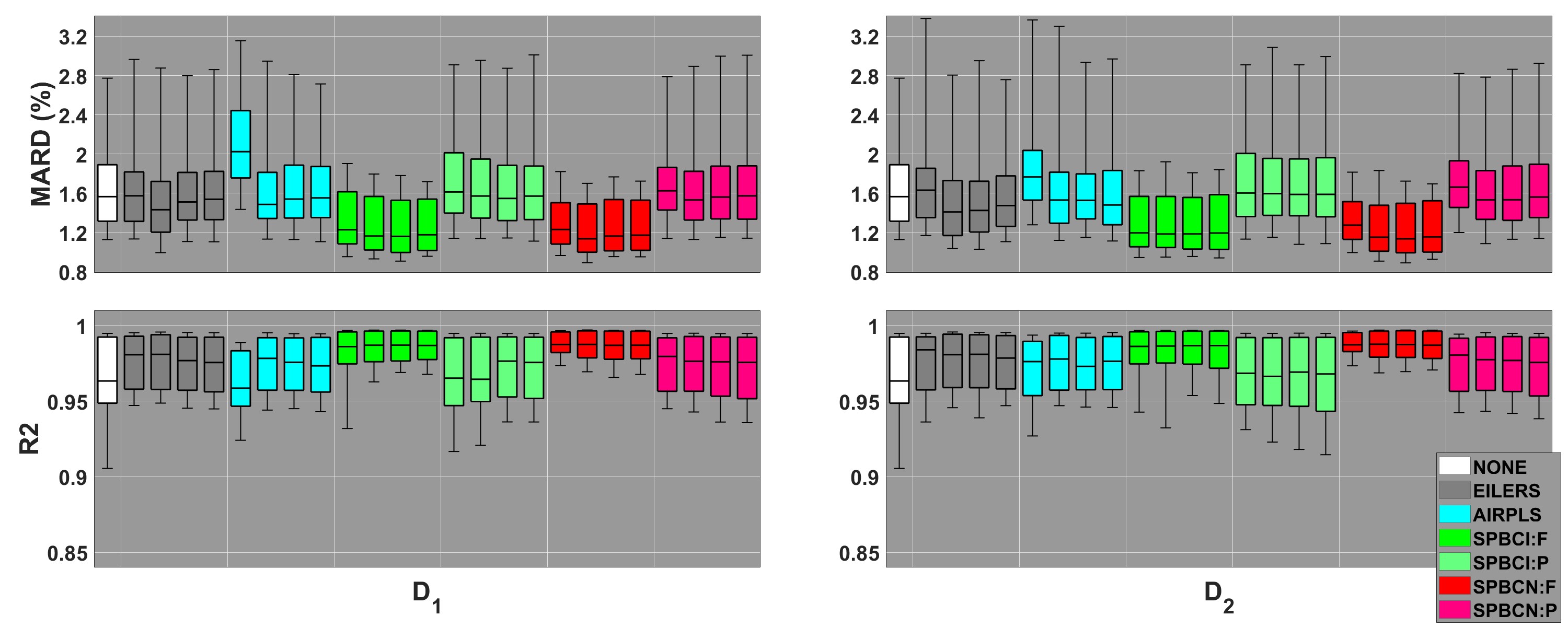}
    \caption{\footnotesize 
    \textbf{Urea ($\bd{y}$) and Fat ($\bd{a}$).}
    Performance across baseline correction methods, and across 200
    data splits.  The first and second columns corresponds to the 
    first and second derivative operators, respectively, while the 
    first and second rows correspond to MARD and R2, respectively.  
    Aside from \NONE, each of the four boxplots associated with the 
    same color correspond (from left-to-right) to 
    $\lambda = \{1,10,100,1000\}$.
    } 
    \label{fig:Milk-Fat-Urea}
\end{figure}
\begin{figure}[th]
    \centering
    \includegraphics[width=6.0in]{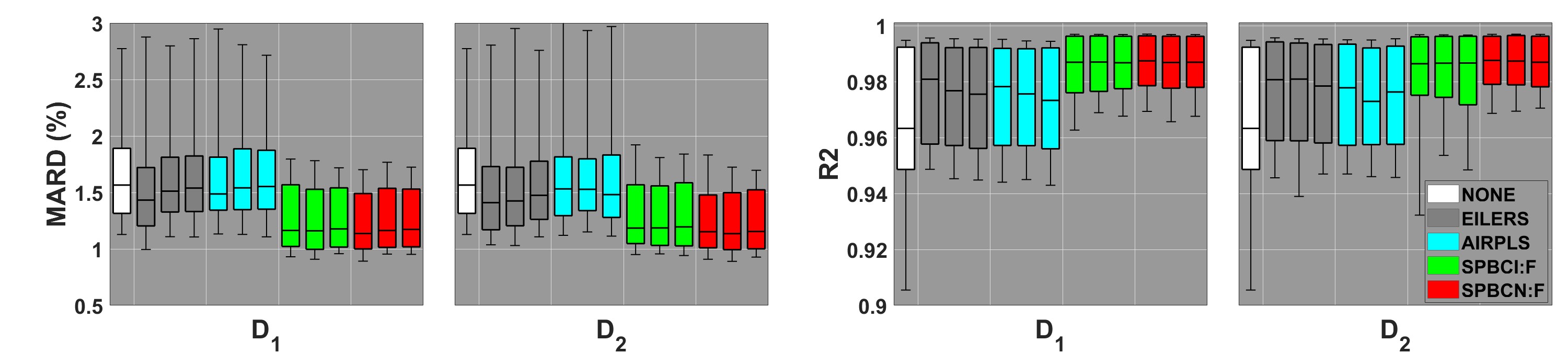}
    \caption{\footnotesize 
    \textbf{Fat ($\bd{a}$) and Urea ($\bd{y}$).}
    Condensed performance display across \NONE, \EILERS, \AIRPLS, 
    \SPBCIF\, and \SPBCNF\, for $\lambda = \{10,100,1000\}$ and 
    across 200 data splits.  The first and second subplots on the 
    left corresponds to MARD while third and fourth subplots 
    correspond to R2.  The first and third columns correspond to 
    the first derivative operator, while the second and fourth 
    columns correspond to the second derivative operator.
    } 
    \label{fig:Milk-Fat-Urea-Condensed}
\end{figure}
We first examine performance where $\bd{a}$ and $\bd{y}$ correspond 
to fat and urea, respectively. Figure (\ref{fig:Milk-Fat-Urea}) 
displays the summary MARD and R2 boxplot performance across all six 
baseline correction methods in addition to \NONE. The first and 
second columns correspond to the first and second derivative 
matrices, while the first and second rows are associated with MARD 
and R2, respectively. Aside from \NONE, each method has four 
boxplots associated with it (all with the same color), and from 
left-to-right, these intra-method boxplots correspond to  
$\lambda = \{1,10,100,1000\}$.  We want to note several archetypal 
patterns of behavior:
\begin{enumerate}
    \item 
    The partial SPBC schemes exhibit poor performance across 
    all $\lambda$ values, and are always non-superior to \NONE.
    \item 
    With respect to intra-method performance, the performance 
    associated with $\lambda=1$, on average, is always non-superior 
    to the boxplots associated with $\lambda = \{10,100,1000\}$.  
    This is especially the case with MARD but less so with R2.
\end{enumerate}
The above performance trends hold not just for urea, but also 
generalize across different analytes and data sets examined in 
this paper.  As a result, and for ease of visualization, we will 
heretofore focus on $\lambda = \{10,100,1000\}$ as well exclude 
the partial SPBC schemes from subsequent consideration.  Figure 
\ref{fig:Milk-Fat-Urea-Condensed} displays the resulting reduced 
set of boxplots, and it is clear that only \SPBCIF\, and \SPBCNF\, 
are superior to the other methods.  Compared to \NONE, the PBC 
approaches of \EILERS\, and \AIRPLS\, exhibit non-inferior 
performance with respect to MARD, but marginally superior R2 
performance.

\subsubsection{Impact of correlation between $\bd{a}$ and $\bd{y}$}
\label{sec:Milk-Correlation}
\begin{figure}[ht]
    \begin{subfigure}[b]{0.91\textwidth}
        \centering
        \includegraphics[width=6.0in]
        {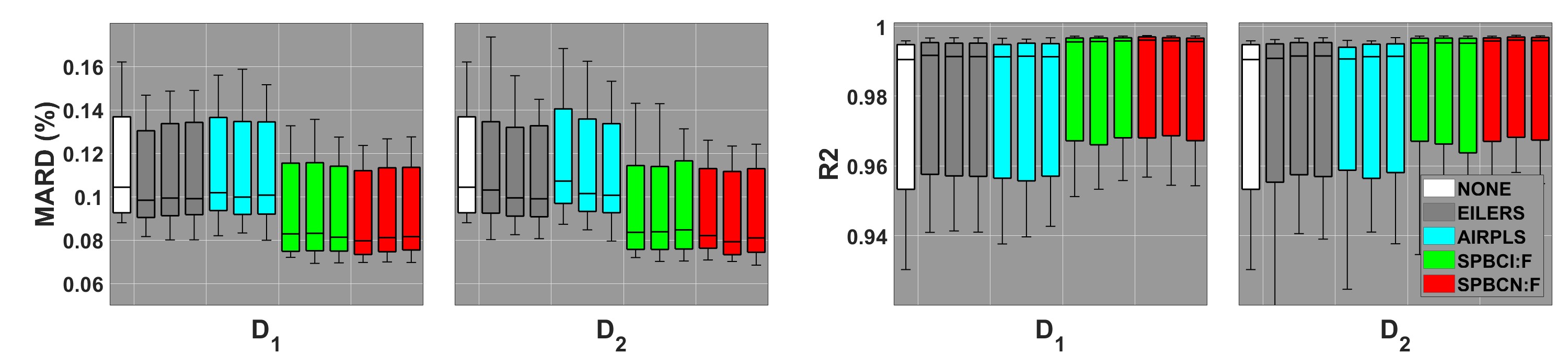}
        \caption{Performance for fat 
        ($\bd{a}$) lactose ($\bd{y}$).}
        \label{fig:Milk-Lactose}
    \end{subfigure}
    \begin{subfigure}[b]{0.91\textwidth}
        \centering
        \includegraphics[width=6.0in]
        {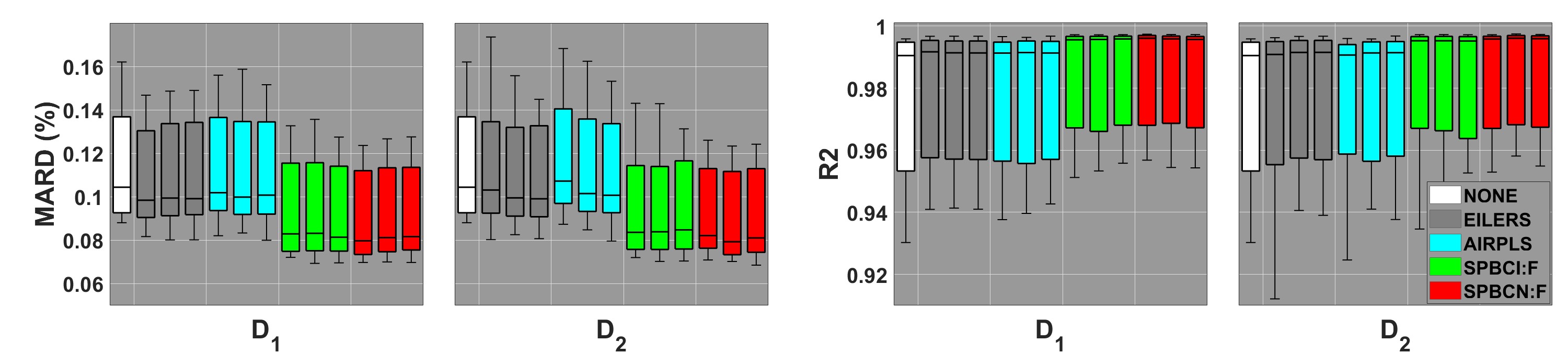}
        \caption{Performance for fat 
        ($\bd{a}$) and protein ($\bd{y}$).}
        \label{fig:Milk-Protein}
    \end{subfigure}
    \begin{subfigure}[b]{0.91\textwidth}
        \centering
        \includegraphics[width=6.0in]
        {Milk-Fat-Urea-Condensed.jpg}
        \caption{Performance for fat 
        ($\bd{a}$) and urea ($\bd{y}$)} 
        \label{fig:Milk-Urea}
    \end{subfigure}
    \begin{subfigure}[b]{0.91\textwidth}
        \centering
        \includegraphics[width=6.0in]
        {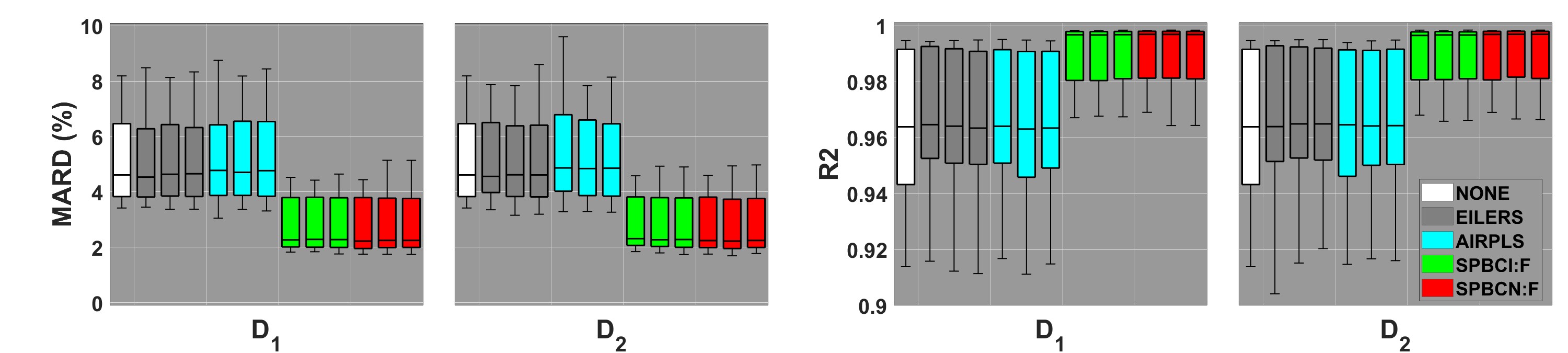}
        \caption{Performance for fat 
        ($\bd{a}$) and solute ($\bd{y}$).}
        \label{fig:Milk-Solute}
    \end{subfigure}
    \begin{subfigure}[b]{0.91\textwidth}
        \centering
        \includegraphics[width=6.0in]
        {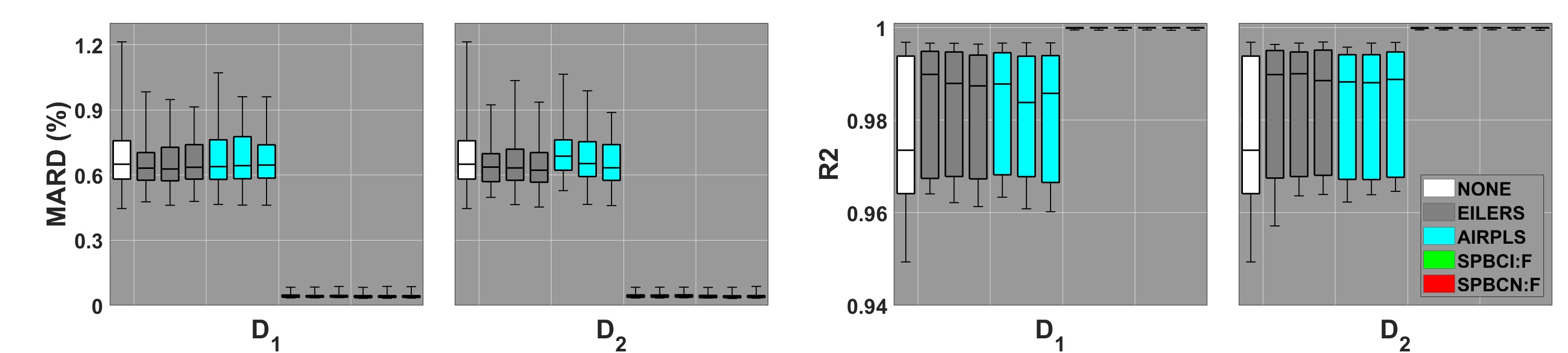}
        \caption{Performance for fat 
        ($\bd{a}$) and dry matter ($\bd{y}$).} 
        \label{fig:Milk-Drymatter}
    \end{subfigure}
    \caption{\footnotesize 
    MARD and R2 performance for the Milk data set using
    instrument NIR-TM3.
    Description-wise, this figure has 
    the same format as Figure \ref{fig:Milk-Fat-Urea-Condensed}.
    The correlations between fat and the other analytes 
    are shown in \ref{tab:milk}.}
    \label{fig:Milk-Correlation}
\end{figure}

\begin{figure}[ht]
    \begin{subfigure}[b]{0.91\textwidth}
        \centering
        \includegraphics[width=6.0in]
        {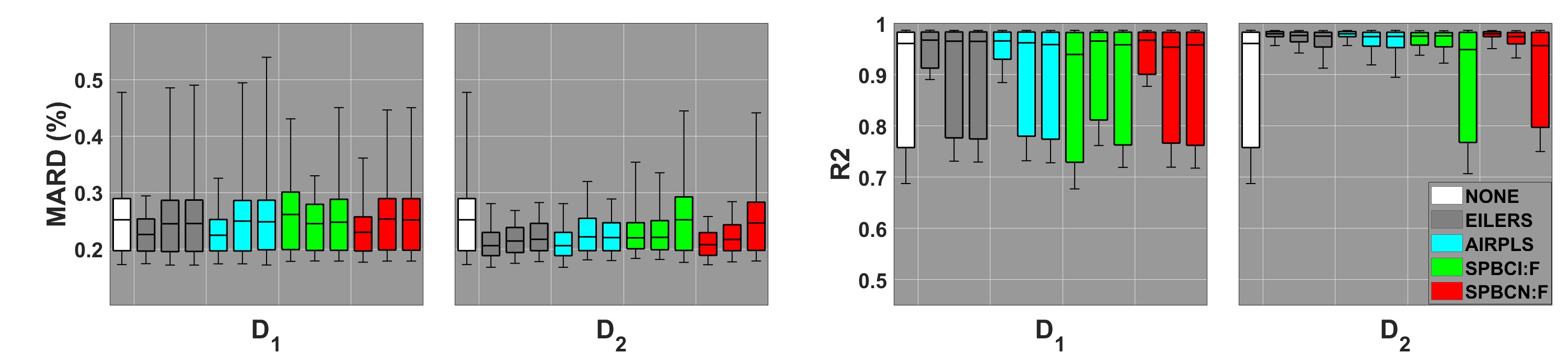}
        \caption{Performance for fat 
        ($\bd{a}$) lactose ($\bd{y}$).}
        \label{fig:Milk2-Lactose}
    \end{subfigure}
    \begin{subfigure}[b]{0.91\textwidth}
        \centering
        \includegraphics[width=6.0in]
        {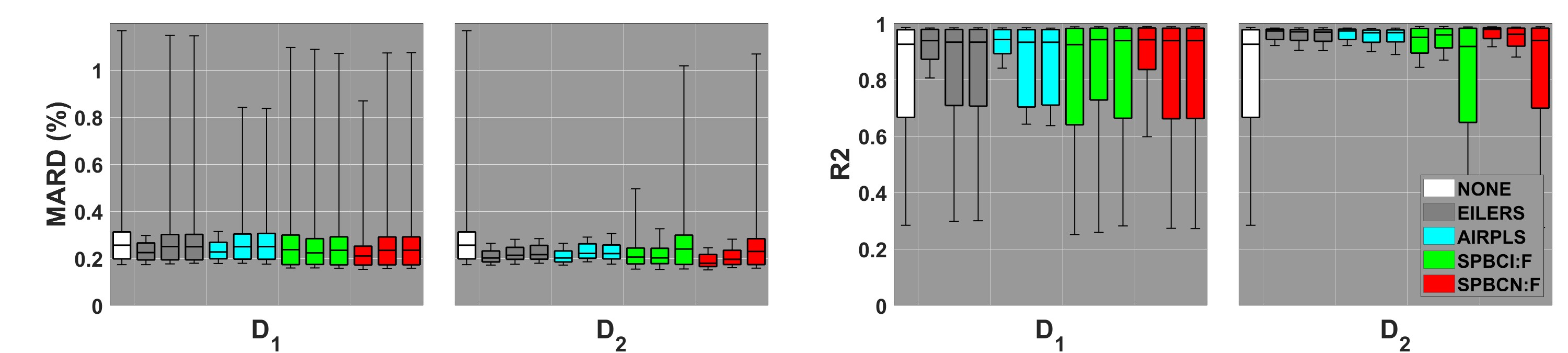}
        \caption{Performance for fat 
        ($\bd{a}$) and protein ($\bd{y}$).}
        \label{fig:Milk2-Protein}
    \end{subfigure}
    \begin{subfigure}[b]{0.91\textwidth}
        \centering
        \includegraphics[width=6.0in]
        {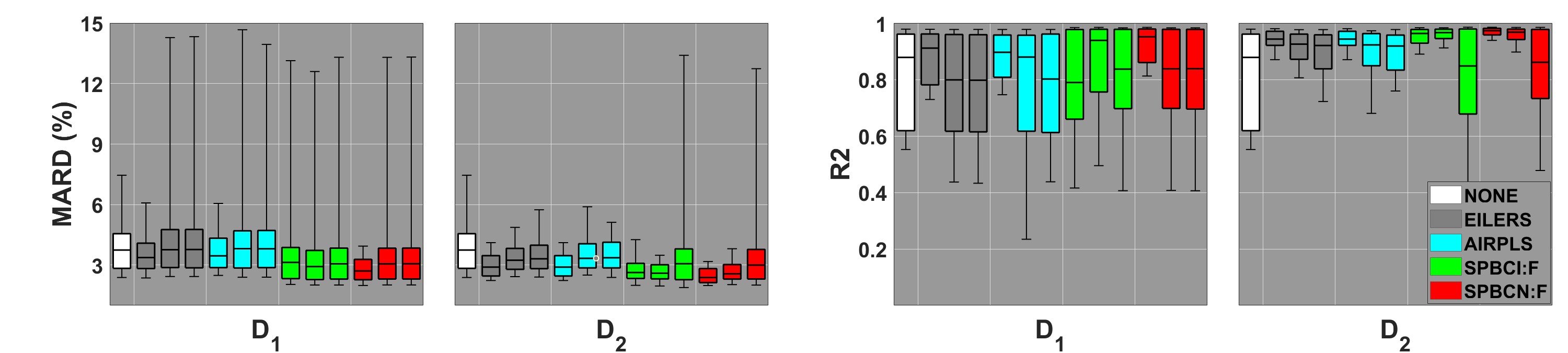}
        \caption{Performance for fat 
        ($\bd{a}$) and urea ($\bd{y}$)} 
        \label{fig:Milk2-Urea}
    \end{subfigure}
    \begin{subfigure}[b]{0.91\textwidth}
        \centering
        \includegraphics[width=6.0in]
        {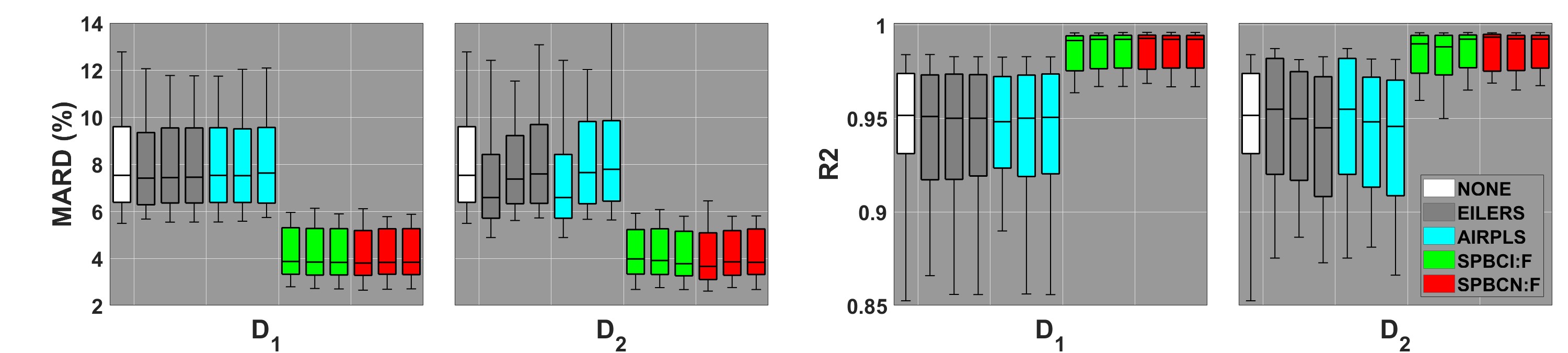}
        \caption{Performance for fat 
        ($\bd{a}$) and solute ($\bd{y}$).}
        \label{fig:Milk2-Solute}
    \end{subfigure}
    \begin{subfigure}[b]{0.91\textwidth}
        \centering
        \includegraphics[width=6.0in]
        {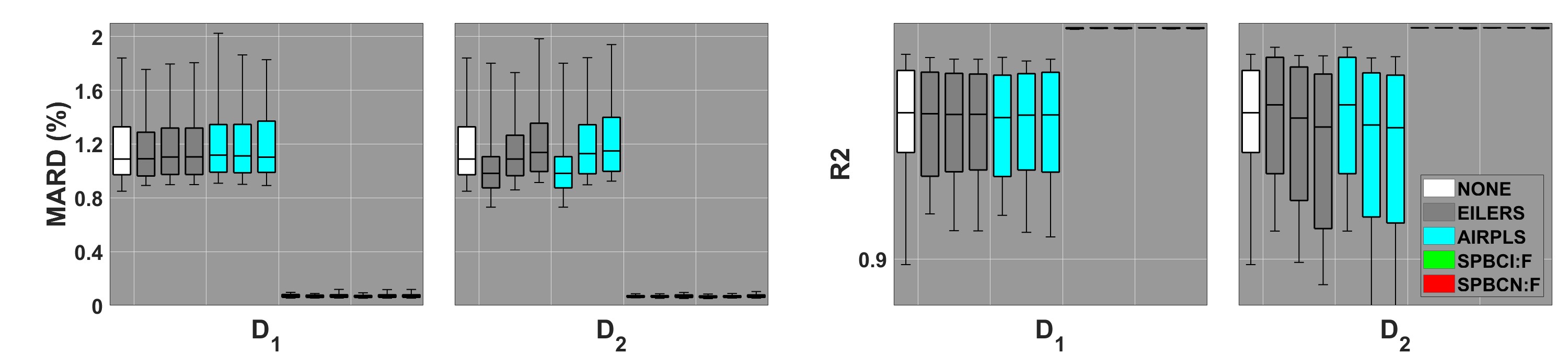}
        \caption{Performance for fat 
        ($\bd{a}$) and dry matter ($\bd{y}$).} 
        \label{fig:Milk2-Drymatter}
    \end{subfigure}
    \caption{\footnotesize 
    The display is the same as 
    \ref{fig:Milk-Correlation}
    except that the performance corresponds 
    to instrument NIR-TM2.
    }
    \label{fig:Milk2-Correlation}
\end{figure}

In this section, we now compare fat ($\bd{a}$) with all the other 
possible analytes $\bd{y}$ (that ones that we want to predict) in 
order of correlation coefficient $r$ magnitude---see Table 
\ref{tab:milk}.
\begin{table}[th]
    \centering
    \begin{tabular}{ |c|c|c|c|c|c|  }
     \hline
     & \multicolumn{5}{|c|}{Correlation Coefficient ($r$) of 
     $\bd{y}$ with fat ($\bd{a}$)} \\
     \hline
     $\bd{y}$ & lactose & protein & urea & solute & dry matter\\
     \hline
     $r$ & 0.1883 & -0.4305 & -0.5480 & 0.7771 & 0.9985\\
    \hline
    \end{tabular}
    \caption{\small Milk data set: The correlation coefficient 
    with fat and each of the other analytes.}
    \label{tab:milk}
\end{table}
Figures \ref{fig:Milk-Correlation} and \ref{fig:Milk2-Correlation}
display MARD and R2 performance across all of these analyte pairs 
for instruments NIR-TM3 and NIR-TM2, respectively.  For the SPBC
approaches, we observe that MARD and R2 performance improves as the 
correlation coefficient magnitude $|r|$ increases.  The improved 
performance with increasing $|r|$ can be explained by examining 
Steps 3 and 4 in Algorithms \ref{alg:SPBCn} and \ref{alg:SPBCi}
(for simplicity of notation, we will drop the subscript $_{(k+1)}$ 
and denote $\bd{Z}_{(k+1)}$ as $\bd{Z}$):
\[
    \begin{array}{rrcll}
    \text{Algorithm 1:} & \bd{Z} & = &  
    \bd{X}(\bd{I} + \lambda^2 \bd{C})^{-1} - \bd{ag}^{\tr}, &
    \bd{g} = (\bd{I} + \lambda^2 \bd{C})^{-1}\bd{w}
    \\[8pt]
    \text{Algorithm 2:} & \bd{Z} & = &  
    \bd{Xwg}^{\tr} - \bd{ag}^{\tr}, &
    \bd{g} = \bd{w}^{\tr}  (\bd{ww}^{\tr} + \lambda^2 \bd{C})^{-1}
    \end{array}
\]
By its very construction, the baseline spectra $\bd{Z}$ is correlated 
with $\bd{a}$, and the baseline-corrected spectra $\bd{X} - \bd{Z}$ 
will likewise be correlated with $\bd{a}$.  If $\bd{a}$ has a strong 
correlation with $\bd{y}$, then the calibration model built from 
$\{\bd{X}_1 - \bd{Z}_{1}, \bd{y}_1\}$ should yield an improved 
prediction for $\bd{y}_2$.  This also explains why the partial 
schemes performed poorly compared to the full scheme.  In the partial 
schemes, we obtain estimates for $\estav$ by building a calibration 
model from $\{\bd{X}_1,\bd{a}_1\}$ and subsequently predicting 
$\bd{a}_2$ from $\bd{X}_2$.  We hope that the prediction will 
be accurate and precise but there is no expectation that the 
prediction estimates will also preserve correlation.  
In effect, the correlation between $\estav$ and $\bd{y}_2$
has been degraded in the partial schemes.

\subsection{Cookie Performance}
\label{sec:CookiePerf}
\begin{figure}[ht]
    \begin{subfigure}[b]{0.91\textwidth}
        \centering
        \includegraphics[width=6.0in]
        {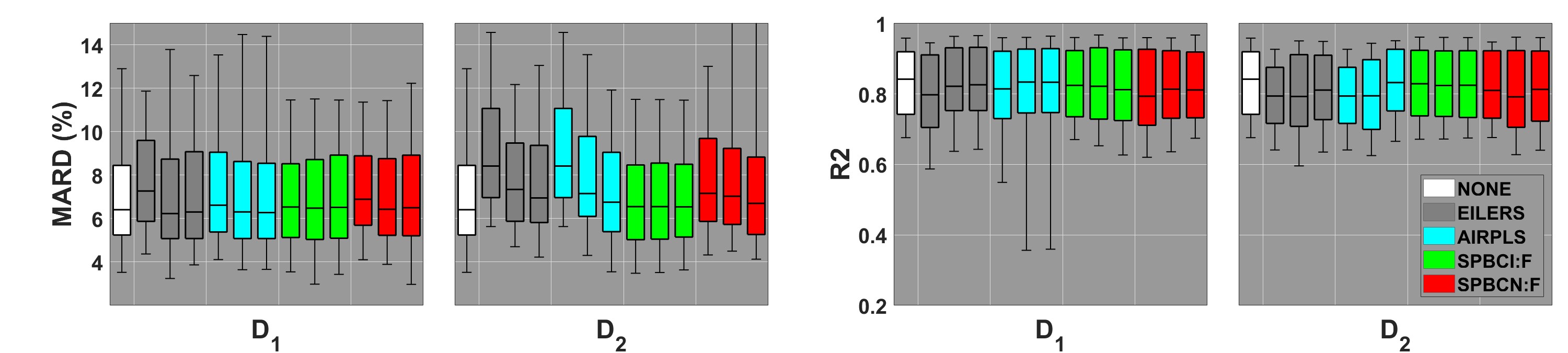}
        \caption{Performance for sucrose 
        ($\bd{y}$) and fat ($\bd{a}$).}
        \label{fig:Cookie-Fat}
    \end{subfigure}
    \begin{subfigure}[b]{0.91\textwidth}
        \centering
        \includegraphics[width=6.0in]
        {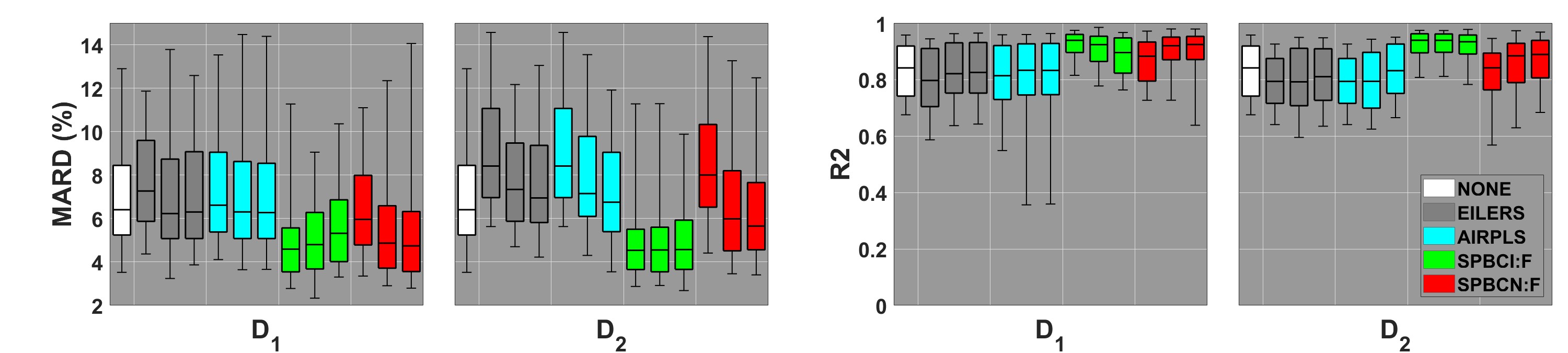}
        \caption{Performance for sucrose 
        ($\bd{y}$) and water ($\bd{a}$).}
        \label{fig:Cookie-Water}
    \end{subfigure}
    \begin{subfigure}[b]{0.91\textwidth}
        \centering
        \includegraphics[width=6.0in]
        {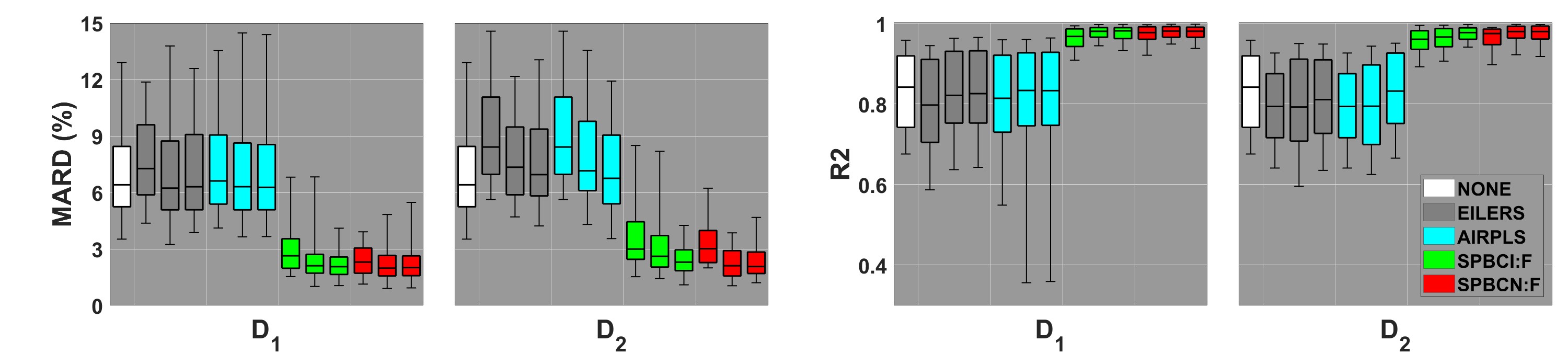}
        \caption{Performance for sucrose 
        ($\bd{y}$) and flour ($\bd{a}$).} 
        \label{fig:Cookie-Flour}
    \end{subfigure}
    \caption{\footnotesize 
    The display is the same as Figure \ref{fig:Milk-Correlation}
    except that the performance corresponds to the Cookie 
    data set and its associated analytes.
    }
    \label{fig:Cookie-Correlation}
\end{figure}

The cookie data set allows us to explore the construction of 
baselines using various analytes as the correlation coefficient 
magnitude between $\bd{y}$ and $\bd{a}$ increases.   Figure 
\ref{fig:Cookie-Correlation} displays performance for three 
pairs of analytes involving sucrose with an increasing degree 
of correlation coefficient magnitude.
\begin{table}[th]
    \centering
    \begin{tabular}{ |C{0.6cm}|C{2.5cm}|C{2.5cm}|C{2.5cm}|  }
     \hline
     & \multicolumn{3}{|c|}{Correlation Coefficient 
     ($r$) of $\bd{a}$ with sucrose $\bd{y}$} \\
     \hline
     $\bd{a}$ & fat & water & flour \\
     \hline
     $r$ & -0.1581 & -0.6860 & -0.9424 \\
    \hline
    \end{tabular}
    \caption{\small Cookie data set: The correlation 
    coefficient between sucrose and each of the other 
    analytes.}
    \label{tab:cookie}
\end{table}
Since the response variable sucrose ($\bd{y}$) is fixed, the 
performance for \NONE\, and the PBC methods of \EILERS\, and 
\AIRPLS\, do not change since the construction of the baselines 
are purely unsupervised---they do not take the analyte $\bd{a}$ 
into account.  As expected, the SPBC performance does change 
(as was the case with the Milk data sets) and this performance 
improves as the correlation coefficient magnitude $|r|$ increases.  
For sucrose ($\bd{y}$) and fat ($\bd{a}$), the analyte pair with 
the lowest correlation coefficient magnitude, none of the baseline 
correction methods outperform \NONE.  With respect to sucrose 
($\bd{y}$) and water ($\bd{a}$), the performance is similar to 
what we observed with the milk data set, i.e., \SPBCIF\, and 
\SPBCNF\, exhibit superior performance compared to \NONE,\, 
\EILERS\, and \AIRPLS.   As with the milk data sets, the analytes 
with the strongest correlation between $\bd{a}$ and $\bd{y}$ yield 
the best performance, particularly with respect to R2.

%
\section{Conclusion and Future Work} 
\label{sec:concl}
%
The SPBC approaches provide a simple extension for estimating 
baselines that incorporate a priori analyte information. There 
are two metaparameters ($\lambda$ and latent dimension) that are 
relatively easy to tune, e.g., MARD and R2 performance were 
observed to be qualitatively invariant across meaningful 
values of $\lambda$ ($\lambda \gg 1$).  SPBC via the full scheme 
provides useful baseline-corrected signals that outperform 
traditional state-of-the-art penalized baseline algorithms 
such as \AIRPLS. 


With respect to the Eilers approach in the case of 
$\bd{D} = \bd{D}_1$, we have developed even faster 
implementations (see Supplement) than Cholesky factorizations.  
In particular, the computation of the singular values and 
loading vectors of $\bd{D}_1$ using closed-form analytical 
formulas are novel in chemometrics.  These fast implementations 
have been socketed into the alternating least squares framework 
of SPBC.  Moreover, the filter factor representations discussed 
in the Supplement allow one to apply SPBC across multiple values 
of $\lambda$ simultaneously.


In this paper, SPBC has only been applied to NIR data sets.
We would like to see if this approach can be applied to other 
spectroscopic modalities such as Raman spectra, fluorescence 
spectra, NMR signals, etc.  The SPBC methods only had superior 
performance for the full scheme, and not for the partial scheme.
We seek to develop alternative partial schemes where better 
estimates for $\estav$ can be obtained.  Alternative schemes 
could include semi-supervised learning where the training data
$\{\bd{X}_1,\bd{a}_1\}$ and $\bd{X}_2$ are used to compute  
$\estav$ (as opposed to just using the $\{\bd{X}_1,\bd{a}_1\}$). 
Improvements in partial scheme development will allow for more 
meaningful use-case scenarios and will lead to more widespread 
adoption.  We have applied SPBC using only one analyte for 
$\bd{a}$.  However, multiple analytes can be accommodated 
into a matrix $\bd{A} = [\bd{a}_1,\ldots,\bd{a}_p]$ such that 
Step 3 in Algorithms \ref{alg:SPBCn} and \ref{alg:SPBCi} can 
be rewritten as $\bd{R} = \bd{X} - \bd{AW}^{\tr}$ and 
$\bd{R} = \bd{XW} - \bd{A}$, respectively.  Moreover, one is 
not necessarily restricted to $\bd{a}$ (or $\bd{A}$) being 
continuously valued reference measurements.  These reference 
measurements could be categorical, and the regression framework 
employed here in this paper could be extended to classification 
algorithms.


\section*{Acknowledgements}
Ramin Nikzad-Langerodi acknowledges funding by the Federal Ministry 
for Climate Action, Environment, Energy, Mobility, Innovation and 
Technology (BMK), the Federal Ministry for Digital and Economic 
Affairs (BMDW), and the State of Upper Austria in the frame of the 
SCCH competence center INTEGRATE [(FFG grant no. 892418)] in the 
COMET - Competence Centers for Excellent Technologies Program 
managed by Austrian Research Promotion Agency FFG, and the FFG 
project zero3 (Grant No. 896399). 

\clearpage
\newpage
\printbibliography[title={Bibliography}]

@book{hansen,
    author = {P.C. Hansen},
    title = {Rank-Deficient and Discrete Ill-Posed Problems},
    publisher = {Society for Industrial and Applied Mathematics}, 
    doi = {10.1137/1.9780898719697},
    url = {https://epubs.siam.org/doi/abs/10.1137/1.978089871969},
    year = {1998}
}

@article{cookbook,
	author = {K.B. Petersen and M.S. Pedersen},
        organization = {Technical University of Denmark},
	title = {The Matrix Cookbook},
	url = {http://www2.imm.dtu.dk/pubdb/pubs/3274-full.html},
	year = {2012}
}

@article{dazzi,
	Author = {A. Dazzi and A. Deniset-Besseau and P. Lasch },
	Journal = {Analyst},
	Pages = {4191-4201},
	Title = {Minimising contributions from scattering 
                 in infrared spectra by means of an 
                 integrating sphere},
	Volume = {138},
        Doi = {10.1039/c3an00381g},
	Year = {2013}
}

@article{ruckstuhl,
	Author = {A.F. Ruckstuhl and M.P. Jacobson and R.W. Field and
                  J.A. Dodd},
	Journal = {Journal of Quantitative Spectroscopy \& 
                   Radiative Transfer},
	Pages = {179-193},
	Title = {Baseline subtraction using robust local regression estimation},
	Volume = {68},
        Doi = {10.1016/S0022-4073(00)00021-2},
	Year = {2001}
}

@article{schecter,
	Author = {I. Schecter},
	Journal = {Analytical Chemistry},
	Pages = {2580-2585},
	Title = {Correction for Nonlinear Fluctuating Background 
                 in Monovariable Analytical Systems},
	Volume = {67},
        Doi = {10.1021/ac00111a014},
	Year = {1995}
}

@article{mazet,
  title = {Background removal from spectra by designing and 
           minimising a non-quadratic cost function},
  author = {Vincent Mazet and C{\'e}dric Carteret and David Brie 
            and J{\'e}r{\^o}me Idier and Bernard Humbert},
  journal = {Chemometrics and Intelligent Laboratory Systems},
  year = {2005},
  volume = {76},
  pages = {121-133},
  doi = {10.1016/J.CHEMOLAB.2004.10.003}
}

@article{Morhac,
  title = {Background removal from spectra by designing and 
           minimising a non-quadratic cost function},
  author = {Vincent Mazet and C{\'e}dric Carteret and David Brie 
            and J{\'e}r{\^o}me Idier and Bernard Humbert},
  journal = {Chemometrics and Intelligent Laboratory Systems},
  year = {2005},
  volume = {76},
  pages = {121-133},
  doi = {10.1016/J.CHEMOLAB.2004.10.003}
}

@article{sg,
  title = {Smoothing and differentiation of data by simplified least 
           squares procedures},
  author = {A. Savitzky and M.J.E. Golay},
  journal = {Analytical Chemistry},
  year = {1964},
  volume = {36},
  issue = {8},
  pages = {1627-1639},
  doi = {10.1021/ac60214a047}
}

@article{msc,
  title = {Linearization and scatter‐correction for near‐infrared reflectance spectra of meat},
  author = {P. Geladi and D. MacDougall and H. Martens},
  journal = {Applied Spectroscopy},
  year = {1985},
  volume = {39},
  issue = {3},
  pages = {491-500},
  doi = {}
}

@article{emsc,
  title = {Extended multiplicative signal correction and spectral interference subtraction: new preprocessing methods for near
  infrared spectroscopy},
  author = {H. Martens and E. Stark},
  journal = {J Pharm Biomed Anal.},
  year = {1991},
  volume = {9},
  issue = {8},
  pages = {625-635},
  doi = {}
}

@article{mancini,
  title = {Study of the scattering effects on NIR data for the prediction
of ash content using EMSC correction factors},
  author = {M. Mancini and Giuseppe Toscano and 
  $\mathring{\text{A}}$smund Rinnan},
  journal = {J Pharm Biomed Anal.},
  year = {1991},
  volume = {9},
  issue = {8},
  pages = {625-635},
  doi = {}
}

@article{schmid,
  title = {Why and How Savitzky-Golay Filters Should Be Replaced},
  author = {M. Schmid and D. Rath and U. Diebold},
  journal = {ACS Measurement Science},
  year = {2022},
  volume = {2},
  pages = {185-196},
  doi = {10.1021/acsmeasuresciau.1c00054}
}

@article{airpls,
	author = {Z.M. Zhang and 
                  S. Chen and 
                  Y.Z Liang},
	journal = {Analyst},
	pages = {1138-1146},
	title = {Baseline correction using adaptive iteratively 
                 reweighted penalized least squares},
	volume = {135},
        issue = {5},
	  doi = {10.1039/b922045c},
	year = {2010}
}

@article{arpls,
	author = {S-J. Baek and 
                  A. Park and 
                  Y-J, Ahn and
                  J. Choo},
	journal = {Analyst},
	pages = {250-257},
	title = {Baseline correction using asymmetrically
                 reweighted penalized least squares smoothing},
	volume = {140},
        issue = {},
	  doi = {10.1039/c4an01061b},
	year = {2015}
}

@article{eilers,
	author = {P.H.C. Eilers},
	journal = {Anal Chem},
	pages = {3631–3636},
	title = {A perfect smoother},
	volume = {75},
        issue = {14},
	  doi = {10.1021/ac034173t},
	year = {2003}
}

@article{eilersals,
	author = {P.H.C. Eilers and H.F.M. Boelens},
	journal = {Report (Leiden University Medical Centre)},
	pages = {},
	title = {Baseline Correction with Asymmetric Least Squares Smoothing},
	volume = {},
        issue = {},
	  doi = {},
	year = {2005}
}

@article{tridiagonal,
	author = {W-C Yueh},
	journal = {Applied Mathematics E-Notes},
	pages = {66-74},
	title = {Eigenvalues of several tridiagonal matrices},
	volume = {5},
        issue = {},
	  doi = {},
	year = {2005},
        url = {https://www.math.nthu.edu.tw/~amen/2005/040903-7.pdf}
}

@article{cookie,
	author = {B.G. Osborne and T. Fearn and A.R. Miller and S. Douglas},
	journal = {Journal of the Science of Food and Agriculture},
	pages = {99-105},
	title = {Application of Near-Infrared Reflectance Spectroscopy 
                 to Compositional Analysis of Biscuits and Biscuit Dough},
	volume = {35},
        issue = {1},
	  doi = {10.1002/jsfa.2740350116},
	year = {1984},
        url = {}
}

@article{milk,
	author = {S. Uusitalo and J. Diaz-Olivares and J. Sumem and
                  E. Hietala and I. Adriaens and W. Saeys and 
                  M. Utriainen and L. Frondelius and M. Pastell and
                  B. Aernouts},
	journal = {Foods},
	pages = {},
	title = {Evaluation of MEMS NIR Spectrometers for On-Farm 
                 Analysis of Raw Milk Composition},
	volume = {10},
        issue = {},
	  doi = {10.3390/foods10112686},
	year = {2021},
        url = {}
}

\pagebreak

\appendix
\section*{Supplement: Numerical Considerations}
\renewcommand{\thesubsection}{\Alph{subsection}}
There are many instances when the most straightforward solution 
of a linear system may not be the most efficient. For example, 
the numerical solution to the linear system 
$\bd{Z} (\bd{I} + \lambda^2 \bd{C}) = \bd{X}$
suggested by \cite{eilers} uses Cholesky factorization on the 
coefficient matrix $\bd{I} + \lambda^2 \bd{C}$ via sparse matrix 
libraries since $\bd{C} = \bd{D}^{\tr} \bd{D}$ is a tridiagonal or 
pentadiagonal matrix if $\bd{D} = \bd{D}_1$ or $\bd{D} = \bd{D}_2$,
respectively.  While computationally sound, there are other 
implementations that are more efficient. Given that these 
system of linear equations are embedded in an alternative least 
squares loop in Algorithms \ref{alg:SPBCn} and \ref{alg:SPBCi}, 
details involving computational speedup are warranted. 

\subsection{Filter factor representation in the Eilers approach}
\label{sec:ffeilers}
Suppose the \emph{reduced} Singular Value Decomposition (SVD) 
of $\bd{D} \in \mathbb{R}^{(n-k) \times n}$ yields 
$\bd{D} = \bd{USV}^{\tr}$ where $\bd{U}$ and $\bd{V}$ 
are orthonormal and 
$\bd{S} = \text{diag}(s_1,s_2,\ldots,s_{n-k})$, $k=\{1,2\}$.
The \emph{full} SVD of $\bd{D}$ similarly yields 
\[
    \bd{D} = 
    [\bd{U},\Unot] 
    \matvec{cc}{\bd{S} & \bd{0} \\ \bd{0} & \bd{0} }
    [\bd{V},\Vnot]^{\tr}
\]
where $\Unot$ and $\Vnot$ are the orthonormal nullspace 
vectors of $\bd{DD}^{\tr}$ and $\bd{D}^{\tr}\bd{D}$, respectively.  
We will only be interested in $\Vnot$ since 
$\bd{C} = \bd{D}^{\tr}\bd{D}$ in Eq.(\ref{eq:EilersNormal}).
Fortunately, the nullspace $\mathcal{N}$ 
of $\bd{D}$ is well characterized \cite{hansen}: 
$\mathcal{N}( \bd{D}_1 ) = \text{span}(\bd{n}_1)$
and 
$\mathcal{N}( \bd{D}_2 ) = \text{span}([\bd{n}_1, \bd{n}_2])$
where
$\bd{n}_1 = \bd{1}_n = [1,1,\ldots,1]^{\tr}
\quad \text{and} \quad 
\bd{n}_2 = [1,2,\ldots,n]^{\tr}$.
Using classical Gram-Schmidt orthogonalization, we obtain 
the orthonormal columns of $\Vnot$:
\begin{equation}
    \voc{1} = 
    {\ds 
    \frac{ 1 }{ \sqrt{n} } \bd{n}_1,
    }
    \quad 
    \voc{2} =  
    {\ds 
    \frac{  \bd{n}_2 - \bd{\mu}_{\bd{n}_2} \bd{n}_1 }
    {\norm{ \bd{n}_2 - \bd{\mu}_{\bd{n}_2} \bd{n}_1 }} 
    },\,\,
    \bd{\mu}_{\bd{n}_2} = \frac{1}{n}(\bd{1}_n^{\tr} \bd{n}_2)
    \label{eq:nullone}
\end{equation}
such that $\Vnot = \voc{1}$ and $\Vnot=[\voc{1},\voc{2}]$ for 
$\bd{D}_1$ and $\bd{D}_2$, respectively.
As a result, we can express the Eliers solution in 
Eq.(\ref{eq:EilersNormal}) as 
\begin{equation}
    \begin{array}{rcl}
    \bd{z} = (\bd{I} + \lambda^2 \bd{C})^{-1} \bd{x} 
    & = & 
    \left( \bd{VV}^{\tr} + \Vnot \Vnot^{\tr} + 
      \lambda^2 \bd{VS}^2 \bd{V}^{\tr} \right)^{-1} \bd{x} 
    \\[8pt]
    & = &  
    \bd{VFV}^{\tr} \bd{x} + \Vnot \Vnot^{\tr} \bd{x}, 
    \quad \bd{F} = \text{diag}(f_1,\ldots,f_{n-k}), \,\, 
    {\ds f_i = \frac{1}{1+\lambda^2 s_i^2} }
    \end{array}
    \label{eq:ffeilers}
\end{equation}    
When $\bd{D} = \bd{D}_1$, then 
$\Vnot \Vnot^{\tr} \bd{x} = 
 \frac{1}{n} \bd{1}_n \bd{1}_n^{\tr} \bd{x} = 
 \bd{1}_n \mu_x$
where $\mu_x = (\bd{1}_n^{\tr} \bd{x})/n$ is the average 
value across the entries of $\bd{x}$. As a result, the 
baseline spectrum $\bd{z}$ can be expressed as a 
linear combination of the loading vectors of $\bd{D}_1$:
\begin{equation}
    \bd{z} = 
    {\ds \sum_{j = 1}^{n-1} c_j \bd{v}_{:j} } +  \mu_x \bd{1}_n, 
    \quad 
    c_j  = \frac{ \bd{v}_{:j}^{\tr} \bd{x} }{ 1+\lambda^2 s_j^2 }.
    \label{eq:EilersExpansion}
\end{equation}
The second term $(\Vnot \Vnot^{\tr})\bd{x}$ in 
Eq.(\ref{eq:ffeilers}) is the fixed or \emph{unregularized 
component} of the solution $\bd{z}$ since the component does 
not depend on $\lambda$.  The diagonal matrix $\bd{F}$ is 
analogous to the filter factor matrix associated with 
standard Tikhonov regularization or ridge regression 
\cite{hansen}.  The contribution of 
the singular vector $\bd{v}_{:j}$ is damped or ``filtered'' 
by its corresponding filter factor $f_j$. 
As $\lambda \rightarrow 0$, $\bd{F} \rightarrow \bd{I}$, 
and the solution $\bd{z}$ approaches $\bd{x}$.  
At the other extreme, as $\lambda \rightarrow \infty$, the 
first term $\bd{VFV}^{\tr} \bd{x}$ in Eq.(\ref{eq:ffeilers}) 
shrinks toward zero and $\bd{z}$ approaches the 
unregularized component $(\Vnot \Vnot^{\tr}) \bd{x}$. 
The SVD-based solution in Eq.(\ref{eq:ffeilers}) also has 
the appealing aspect in that the solution can be vectorized 
across multiple values of $\lambda$. 
Next we will discuss how the loading vectors $\bd{v}_{:j}$ and 
singular values $s_i$ of $\bd{D}_1$ can be computed without the 
need of the SVD.

\subsection{The eigenstructure of $\bd{D} = \bd{D}_1$}
\label{sec:eigen}
One can exploit the tridiagonal structure of  
$\bd{C} = \bd{D}_1^{\tr}\bd{D}_1 = \bd{VS}^2\bd{V}^{\tr}$ 
to compute the singular values and loading vectors \emph{without 
the need of the SVD}.  We first note that matrix $\bd{C}$ is of
a tridiagonal form 
\begin{equation}
    \bd{D}_1 = \matvec{cccccccc}
    { b-d    & c      & 0      & 0      & \cdots & 
      0      & 0      & 0 \\
      a      & b      & c      & 0      & \cdots & 
      0      & 0      & 0 \\
      0      & a      & b      & c      & \cdots & 
      0      & 0      & 0 \\ 
      \cdots & \cdots & \cdots & \cdots & \cdots & 
      \cdots & \cdots & \cdots \\ 
      0      & 0      & 0      & 0      & \cdots & 
      a      & b      & c \\
      0      & 0      & 0      & 0      & \cdots & 
      0      & a      & b-d }
    \in \mathbb{R}^{n \times n}
    \label{eq:tri}
\end{equation}
where $a=c=-1$, $b=2$ and $d=1$.  The near Toeplitz-like 
structure (Topelitz matrices are banded matrices with 
constant diagonal elements) of Eq.(\ref{eq:tri}) allows the 
singular values $s_j$ and loading vectors 
$\bd{v}_{:j} = [v_{1j},\ldots,v_{nj}]^{\tr}$ of $\bd{D}_1$
to be analytically constructed using symbolic 
calculus\cite{tridiagonal}:
\[
    s_j^2 = 2 - 2\cos\left( \frac{(n-j)\pi}{n} \right), \,\,
    v_{ij} = \cos\left( \frac{(n-j)(2i-1)\pi}{2n} \right),\,\,
    i,j = 1,2,\ldots,n.
\]
This exploitation of the near-Toeplitz structure 
of $\bd{C} = \bd{D}_1^{\tr} \bd{D}_1$ is novel in baseline correction.

\begin{figure}[th]
    \begin{subfigure}[b]{0.95\textwidth}
        \centering
        \includegraphics[width=6.5in]{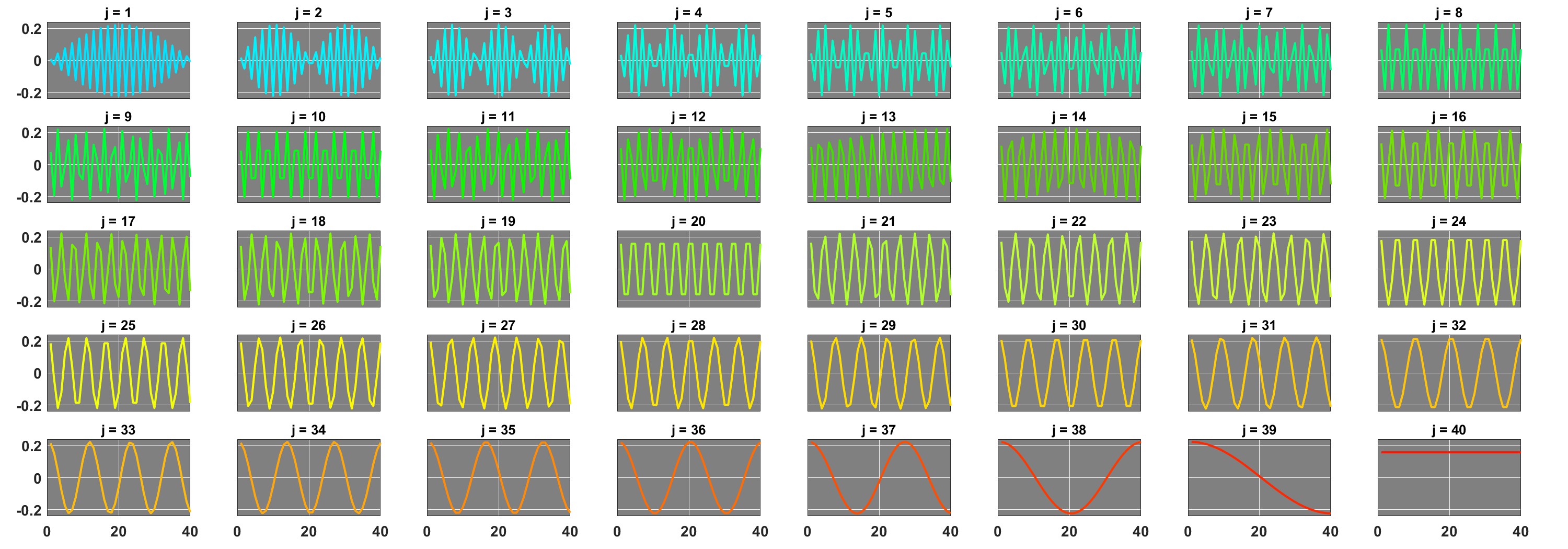}
        \caption{\footnotesize The loading vectors 
        $\bd{v}_{:j} = [v_{1j},v_{2j},\ldots,v_{40j}]^{\tr}$ 
        associated with $\bd{D}_1 \in \mathbb{R}^{39 \times 40}$ are 
        displayed.  For each subplot, the $y$-axis corresponds to 
        the value of $v_{ij}$ while the $x$-axis corresponds to 
        $i=\{1,2,\ldots,40\}$.}
        \label{fig:loadings}
    \end{subfigure}
    \begin{subfigure}[b]{0.40\textwidth}
        \centering
        \includegraphics[width=3.0in]{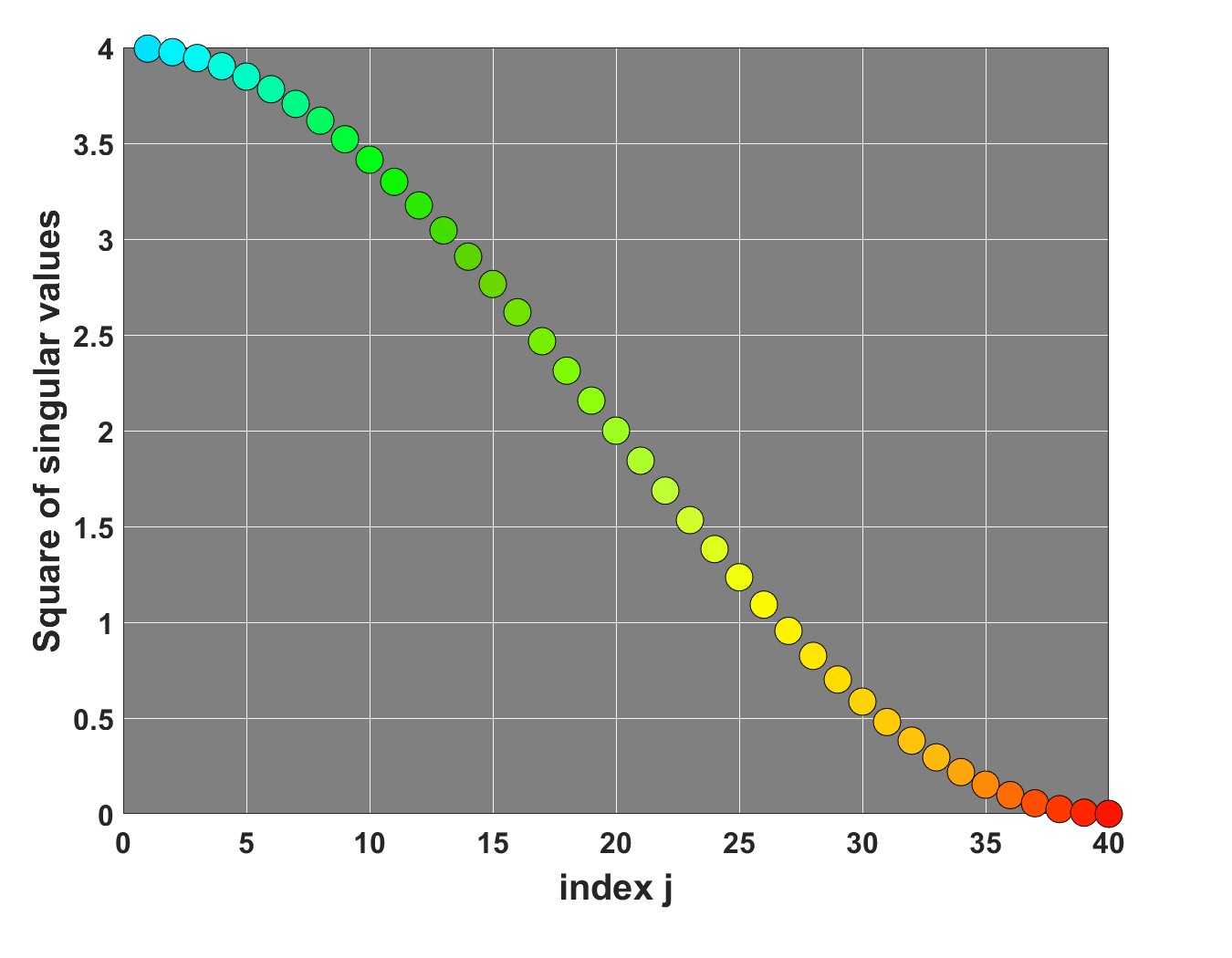}
        \caption{\footnotesize The singular values $s_j^2$
        plotted as a function of index $j$.}
        \label{fig:singular}
    \end{subfigure}
    \hfill
    \begin{subfigure}[b]{0.5\textwidth}
        \centering
        \includegraphics[width=3.32in]{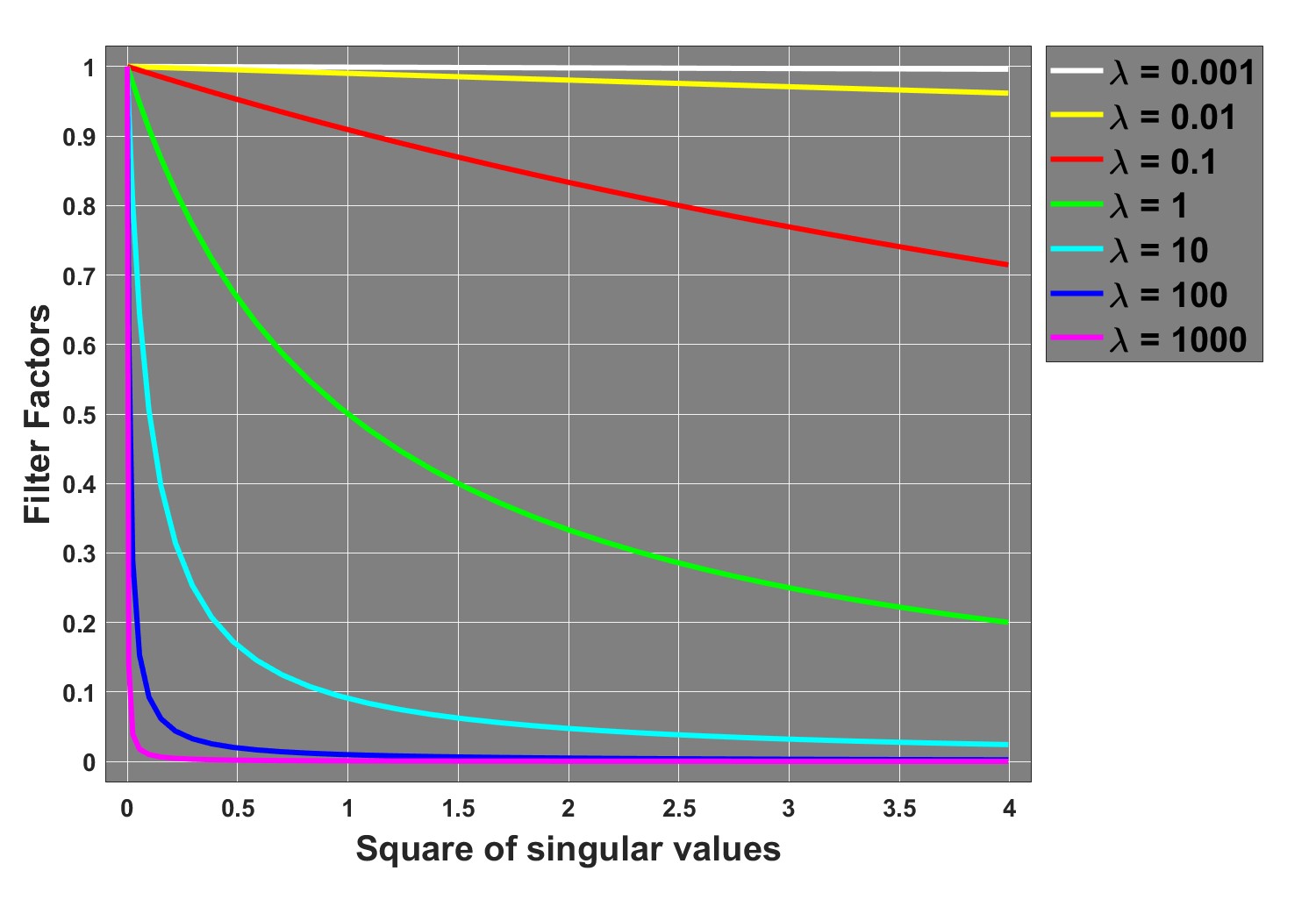}
        \caption{\footnotesize The filter factors 
        $f_j = \frac{1}{1+\lambda^2 s_j^2}$
        plotted as a function of $s_j^2$.}
        \label{fig:filter}
    \end{subfigure}
    \caption{\footnotesize 
    The loadings vectors, singular and filter factors are displayed 
    for a first derivative operator matrix 
    $\bd{D}_1 \in \mathbb{R}^{39 \times 40}$.}
    \label{fig:EigenStruct}
\end{figure}

To illustrate the eigenstructure of the derivative operator, we 
compute the analytical-based SVD of 
$\bd{C}=\bd{D}_1^{\tr}\bd{D} = \bd{VS}^{2}\bd{V}^{\tr}$ 
where $n=40$ such that 
$\bd{D}_1 = \bd{USV}^{\tr} \in \mathrm{R}^{39 \times 40}$.
Figure \ref{fig:loadings} shows the loading vectors $\bd{v}_{:j}$ 
in $\bd{V} = [\bd{v}_{:1},\bd{v}_{:,2},\ldots,\bd{v}_{:40}]$
while Figure \ref{fig:singular} displays the square of the 
singular values $\{s_1^2,s_2^2,\ldots,s_{40}^2\}$. The last 
loading vector $\bd{v}_{40}$ actually corresponds to the 
nullspace vector $\voc{1} = \frac{1}{\sqrt{n}} \bd{1}_{n}$. 
Filter \ref{fig:filter} displays the value of the filter 
factors $f_j = 1/(1+\lambda^2 s_j^2)$ for 
$\lambda=\{0.001,0.01,0.1,1,0.10,100,1000\}$---each 
colored curve 
corresponds to a different value of $\lambda$.  The loading 
vector $\bd{v}_{:j}$ and singular value $s_j^2$ associated 
with each index value of $j$ has its own color: as $j$ 
increases in value, the colors vary from blue (high frequency) 
to red (low frequency). Compared to most matrices, the loading 
vectors of $\bd{D}_1$ (and $\bd{D}_2$ as well) are unusual in 
that the number of sign changes (the number of times $v_{ij}$ 
crosses the $x$-axis) \emph{decreases} as $j$ increases.
The filter factor curves indicate that the terms in 
Eq.(\ref{eq:EilersExpansion}) associated with high frequency 
loading vectors (the blues and greens) are easily 
damped by moderately large values of $\lambda$, whereas
the low frequency loading vectors are preserved except for 
the largest values of $\lambda$.

\subsection{Filter Factor Representations in \SPBCN\, and \SPBCI}
\label{sec:ffspbc}
For \SPBCN, Step 4 of Algorithm \ref{alg:SPBCn}, i.e., 
$\bd{Z}_{(k+1)} = \bd{R} (\bd{I} + \lambda^2 \bd{C})$
(where $\bd{R} = \bd{X} - \bd{aw}^{\tr}$)
is the computational bottleneck of the alternating least squares 
procedure.  Its solution is the same as Eq.(\ref{eq:ffeilers}) 
except that the matrix $\bd{X}$ is replaced with $\bd{R}$:
\begin{equation}
    \bd{Z}_{(k+1)}
    =  
    \bd{R} \left( \bd{VFV}^{\tr} + \Vnot \Vnot^{\tr} \right)
    \bd{RVFV}^{\tr} + \bd{R} \Vnot \Vnot^{\tr},  
\label{eq:spbc_soln}
\end{equation}
For \SPBCI, 
let $\bd{X} = \bd{P \Sigma Q}^{\tr}$ be the \emph{reduced} Singular 
Value Decomposition (SVD) of $\bd{X}$ where $\bd{P}$ and $\bd{Q}$ 
are orthonormal and 
$\bd{\Sigma} = \text{diag}(\sigma_1,\sigma_2,\ldots,\sigma_r)$
where $r$ is the rank $\bd{X}$.  Similarly, let 
\[
\bd{X} = 
[\bd{P} \quad \bd{P}_0]  
\matvec{cc}{ \bd{\Sigma} & \bd{0} \\
               \bd{0}     & \bd{0} }
[\bd{Q} \quad \bd{Q}_0]^{\tr}               
\]
be the \emph{full} SVD where $\text{span}(\bd{Q}_0)$ 
is the nullspace of $\bd{X}$.  In Step 2 of Algorithm 
$\ref{alg:SPBCi}$, the linear system 
$\bd{B}^{\tr}\bd{Bw} = \bd{B}^{\tr} \bd{a}$ 
where $\bd{B} = \bd{X} - \bd{Z}_{(k)}$ 
can then be rewritten as 
\begin{equation}
    \bd{X}^{\tr} \bd{Xw} =  \bd{d}_{(k)}, 
    \quad
    \bd{d}_{(k)} = \bd{X}^{\tr} \bd{a} + 
    \bd{B}^{\tr} \bd{Z}_{(k)} \bd{w} + 
    \bd{Z}_{(k)}^{\tr}(\bd{Xw} - \bd{a})
    \label{eq:svdsoln_spbci}
\end{equation}
In this case, the coefficient matrix $\bd{X}^{\tr} \bd{X}$
on the left-hand-size is constant, and as a result, the 
solution can be expressed using the basis vectors in 
$\bd{Q}$ and $\bd{Q}_0$.  Due to the high correlation 
of spectra in $\bd{X}$, instead of solving 
$\bd{X}^{\tr} \bd{Xw} = \bd{d}_{(k)}$ 
in Step 2 of Algorithm \ref{alg:SPBCi}, we will instead solve 
$( \bd{X}^{\tr}\bd{X}  + \tau^2 \bd{I} ) \bd{w} = \bd{d}_{(k)}$
via ridge regression.\footnote{The ridge parameter will be 
intentionally chosen to be small to ensure numerical stability.  
We will not try to optimize $\tau>0$ as a tuning parameter.}
As result, the solution $\bd{w}$ in Step 2 can be written as
\[
    \bd{w} = 
    \bd{QFQ}^{\tr} \bd{d}_{(k)}  + 
    \bd{Q}_0 \bd{Q}_0^{\tr} \bd{d}_{(k)}
    \quad \text{where} \quad
    \bd{F} = \text{diag}(f_1,\ldots,f_r), \,\,
    f_i = \frac{1}{\sigma_i^2 + \tau^2}
\]
If $\bd{X}$ has full column rank, i.e., $m \geq n$ and $r=n$, 
then $\bd{Q}_0$ will empty and the solution can be written as
$\bd{w} = \bd{QFQ}^{\tr} \bd{d}_{(k)}$.
Since the ridge regression occurs within an alternative least 
squares loop, it is prudent to 
compute the SVD of $\bd{X}$ once at the very beginning of the loop, and 
then re-use the pre-computed SVD components (the singular 
values in $\bd{\Sigma}$, and the loading vectors in $\bd{Q}$ 
and the nullspace vectors in $\bd{Q}_0$) over-and-over again.

\end{document}